\documentclass[runningheads]{llncs}
\usepackage{tikz}
\usepackage{comment}
\usepackage{amsmath,amssymb} % define this before the line numbering.
\usepackage{color}
\usepackage{kotex}
\usepackage{CJKutf8}
\usepackage{makecell, multirow}   % Table Line Break
\usepackage{pifont}     % Table OX     
\usepackage{xcolor}     % Table OX
\usepackage{pbox}       % Table alignment
\usepackage{booktabs}   % Table (\toprule, \cmidrule)
\usepackage{subcaption}
\usepackage{xspace}
\usepackage{hyperref}
\usepackage{graphicx} 
\usepackage[accsupp]{axessibility}  % Improves PDF readability for those with disabilities.

\usepackage{algorithm2e}
\usepackage{physics}    % Norm operation

\newcommand{\cmark}{\textcolor{green!80!black}{\ding{51}}}
\newcommand{\xmark}{\textcolor{red}{\ding{55}}}
\newcommand{\ourdata}{AnimeCeleb\xspace}
\newcommand{\model}{\emph{AniMo}\xspace}

\mainmatter
\def\ECCVSubNumber{6068}  % Insert your submission number here

\title{AnimeCeleb: Large-Scale Animation \\ CelebHeads Dataset for Head Reenactment}

\titlerunning{AnimeCeleb: Large-Scale Animation CelebHeads Dataset}
\author{Kangyeol Kim\inst{*1,4} \and
Sunghyun Park\inst{*1} \and
Jaeseong Lee\inst{*1} \and \\
Sunghyo Chung\inst{2} \and
Junsoo Lee\inst{3} \and
Jaegul Choo\inst{1,4}}

\authorrunning{K. Kim et al.}

\institute{KAIST \and Korea University \and Naver Webtoon \and Letsur Inc. \\
\email{\{kangyeolk, psh01087, wintermad1245, jchoo\}@kaist.ac.kr \\
s94021@korea.ac.kr \tt junsoolee93@webtoonscorp.com} \\
* indicates equal contributions.}

\begin{document}

\maketitle

\begin{center}
    \centering
    \vspace{-0.4cm}
    \includegraphics[width=\linewidth]{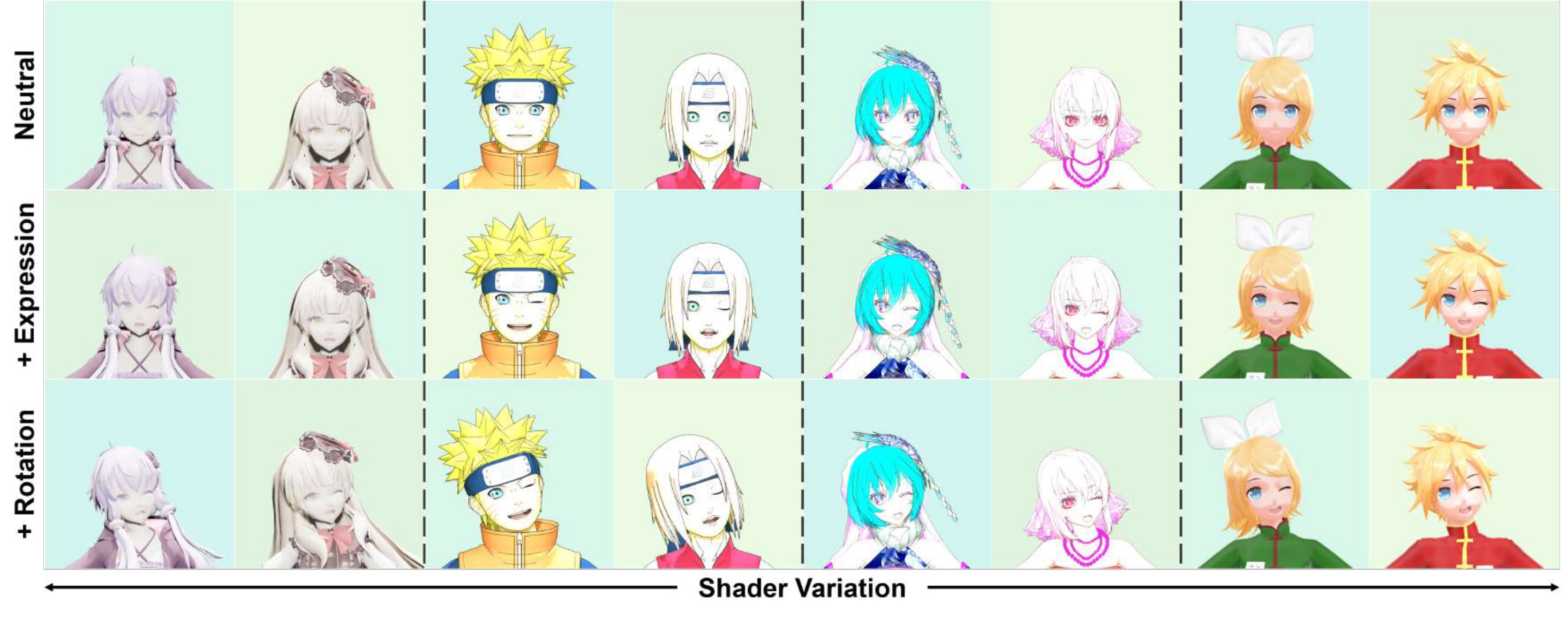}
    \vspace{-0.9cm}
    \captionof{figure}{(Better viewed in color) Examples of our AnimeCeleb. Including a canonical head (\textit{Neutral}), the AnimeCeleb contains expression-changed (+\textit{Expression}) and head-rotated (+\textit{Rotation}) images with varying shaders.}
    \label{fig:teaser}
    \vspace{-0.2cm}
\end{center}

%%%%%%%%% ABSTRACT
\begin{abstract}
    We present a novel Animation CelebHeads dataset (AnimeCeleb) to address an animation head reenactment.
    Different from previous animation head datasets, we utilize a 3D animation models as the controllable image samplers, which can provide a large amount of head images with their corresponding detailed pose annotations. 
    To facilitate a data creation process, we build a semi-automatic pipeline leveraging an open 3D computer graphics software with a developed annotation system.
    After training with the AnimeCeleb, recent head reenactment models produce high-quality animation head reenactment results, which are not achievable with existing datasets.
    Furthermore, motivated by metaverse application, we propose a novel pose mapping method and architecture to tackle   a cross-domain head reenactment task.
    During inference, a user can easily transfer one's motion to an arbitrary animation head.
    Experiments demonstrate an usefulness of the AnimeCeleb to train animation head reenactment models, and the superiority of our cross-domain head reenactment model compared to state-of-the-art methods.
    Our dataset and code are available at \href{https://github.com/kangyeolk/AnimeCeleb}{\textit{this url}}.
\end{abstract}

\vspace{-0.7cm}

%%%%%%%%% BODY TEXT
\section{Introduction}

Recent head reenactment methods~\cite{latentpose,ren2021pirenderer,firstorder} show impressive results on controlling a human head motion after trained with large-scale human talking head video datasets~\cite{chung2018voxceleb2,nagrani2017voxceleb}.
The common approaches~\cite{fewadvneural,fastbilayer,firstorder,ren2021pirenderer} for this task is to learn diverse motion changes between two contiguous frames, which require a large amount of head videos to train a high-performing neural network model.
Due to the dependency of human video datasets, such approaches show weak generalization capacity on the animation domain, because animation characters have distinct appearances (\textit{e.g.}, explicit lines and large eyes) compared to the human head ones.
Our key contribution is to construct a large-scale animation head dataset, AnimeCeleb, for head reenactment, which deems as a data-centric solution to produce high-quality reenactment results on the animation domain.

Obviously, a standard approach to build an animation dataset would be to collect the images from comic books and cartoon films.
Instead, we propose a principled manner to construct animation dataset, where 3D animation models serve as valuable image samplers.
This leads to three following benefits.
First, we can ceaselessly simulate the specified pose\footnote{Throughout this paper, we mean by the 'pose' the information about head rotation, translation, and facial expression.} of a 3D animation model, enabling to generate an \textit{unlimited} number of multi-pose images of the same identity.
Second, the simulated poses are easily obtainable as detailed pose vectors, where each dimension represents an individual semantic of an expression or a head angle.
Lastly, a 3D vector graphics environment gives freedom to render the \textit{arbitrary} resolution images with various shaders (See Fig.~\ref{fig:teaser} horizontal axis). 
These strengths bring multiple use cases including the animation head reenactment and intuitive pose editing.

Technically, our data creation process involves 3D animation model collection, semantic annotation and image rendering.
In this process, we first collect the 3D animation models spanning a wide range of animation characters.
The collected 3D models contain a set of morphs that can deform appearances of the 3D models in face and body part.
To identify suitable morphs relevant to the head reenactment task, we develop an annotation system to filter the expression-irrelevant morphs.
We employ Blender\footnote{https://www.blender.org/} that can execute codes for a head detection and a pose manipulation to enable an automatic image rendering.

A great interest of an animation domain is to transfer a user's motion to the animation character, which is potentially applicable in a metaverse and a virtual avatar system.
In this paper, we focus on transferring a user's pose to the animation character, and refer to this problem as a \textit{cross-domain head reenactment task}.
A plausible solution to the task is building a shared pose representation space across the domains (\textit{i.e.}, human and animation).
We use 3D morphable model (3DMM) parameters as the shared pose representation, which is widely used in recent numerous head reenactment studies~\cite{gafni2021dynamic,zhang2021facial,ren2021pirenderer,wang2021cross,guo2021ad}.
3DMM is a parametric face modeling method that provides powerful tools for describing human heads with semantic parameters.
Since the AnimeCeleb pose vector is not compatible with 3DMM, we newly propose a \textit{pose-mapping} method to transform an AnimeCeleb pose vector to 3DMM parameters.
To be specific, we compute a set of distinct 3DMM parameters to describe the semantics that the AnimeCeleb includes, and combine it to obtain 3DMM parameters corresponding to a AnimeCeleb pose vector.
Owing to the pose mapping, we can guarantee that both the AnimeCeleb and VoxCeleb~\cite{nagrani2017voxceleb}, a human head video dataset, share the pose representations.
Furthermore, we propose a new architecture called an animation motion model (\model), in which datasets from different domains are used to learn how to manipulate a head image according to the motion residing in the shared representations.
In this manner, our model is capable of transferring a human head motion represented as 3DMM parameters to an animation head.\footnote{Related work regarding to the AnimeCeleb and the proposed algorithm is provided in supplementary material.}

In summary, our contributions to animation research are as follows:
\vspace{-0.1cm}
\begin{itemize}
    \item We propose a \emph{novel data creation pipeline} and present a \emph{public large-scale animation head dataset} \ourdata, which contains groups of high-quality images and their corresponding pose vectors.
    \item We newly propose a \textit{pose-mapping} method and a cross-domain head reenactment model \model, which jointly lead to a seamless motion transfer from a \textbf{human head} to an \textbf{animations head}.
    \item We demonstrate the effectiveness of \ourdata in training head reenactment baselines, and experimental results show the superiority of \model on cross-domain head reenactment compared to state-of-the-art methods. 
\end{itemize}

\vspace{-0.4cm}
\section{Animation CelebHeads Dataset}\label{animeceleb}
\vspace{-0.1cm}
We first describe each step of the data creation of the AnimeCeleb in Section~\ref{data_generation_process}.
Next, AnimeCeleb properties and statistics are given in Section~\ref{dataset_description}.
In Section~\ref{animation_head_reenactment}, we show the animation head reenactment results on the AnimeCeleb and other animation datasets.

\vspace{-0.3cm}
\subsection{Data Creation Process}\label{data_generation_process}
\begin{figure}[t!]
    \centering
    \includegraphics[width=1.0\linewidth]{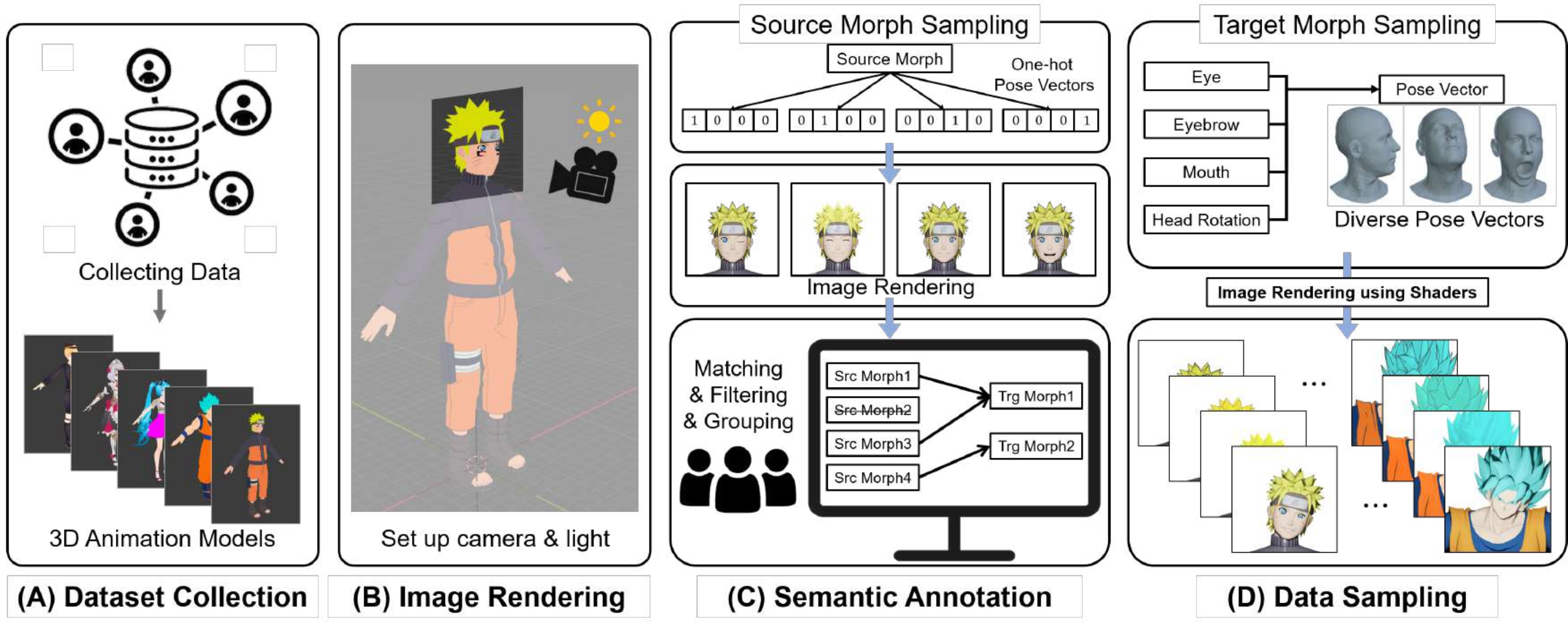}
    \vspace{-0.65cm}
    \caption{\textbf{Dataset Creation Pipeline Overview.} 
    3D animation models are collected from two different websites (A). 
    Then, a head part of the collected model are rendered after applying a morph with maximum intensity (B); these are then used for semantic annotation (C). 
    In a data sampling step, sampled target morphs are used to compose pose vectors that serve as conditions to produce multi-pose images with diverse facial expressions and head rotations.}
    \vspace{-0.75cm}
    \label{fig:pipeline}
\end{figure}

Fig.~\ref{fig:pipeline} depicts the overall process of the data creation pipeline.
In the following, we provide details of each step from (A) to (D).

\noindent\textbf{Data Collection (A).}
We collected 3D animation models from two different web sites: DevianArt\footnote{https://www.deviantart.com/} and Niconi solid\footnote{https://3d.nicovideo.jp/}.
Since all 3D animation models are copyrighted by their creators, we carefully confirmed the scope of rights and obtained permission from reachable authors.
Finally, we acquired 3613 usable 3D animation models in total. 
We will release all 3D animation model \textit{artists' list} along with the AnimeCeleb to acknowledge the credits of the artists.

The collected 3D animation models contain two essential components.
The first component is the \textbf{morphs} that can alter appearances of a 3D animation model on face or body parts. 
We are able to change an individual morph's continuous value ranging from [0, 1], and obtain a transformed appearance of a 3D animation model; for example, an animation head with open mouth in 0.3 proportion can be generated. 
The second one is the \textbf{bones} that can control head angles (\textit{i.e.}, yaw, pitch and roll axes).
In specific, the head angles are controlled by applying a rotation matrix to the neck bone.

% Figure - Dataset Generation Precedure

\noindent\textbf{Image Rendering (B).}
To achieve an automatic sampling using 3D animation models, we develop a 2D head image creation pipeline built on Blender: an open source 3D computer graphics software that supports the visualization, manipulation and rendering of 3D animation models.
To successfully render the animation head images in Blender, we need to consider three aspects: (1) camera position, (2) light condition, and (3) image resolution.

We set the camera position based on a neck bone position with the aim of capturing the head part.
In respect to the light condition, we use a directional light point along the negative \textbf{y}-axis: frontal direction of an animation character (See Fig.~\ref{fig:pipeline} (B)).
Before rendering, we set the resolution of the images as $256 \times 256$, which is a standard resolution used in previous head reenactment methods~\cite{firstorder,ren2021pirenderer}.
Nonetheless, since the AnimeCeleb images are rendered from a 3D vector graphics model, we can create a higher image resolution (\textit{e.g.}, $1024 \times 1024$).
To demonstrate its extensive usage, we present various generated samples under different conditions in the supplementary material.
Note that the rendered images contain an alpha channel as a transparent background, which can separate the foreground animation character and the background.

% Figure - Morph Samples & Morph Distribution
\begin{figure}[t!]
    \centering
    \includegraphics[width=1.0\linewidth]{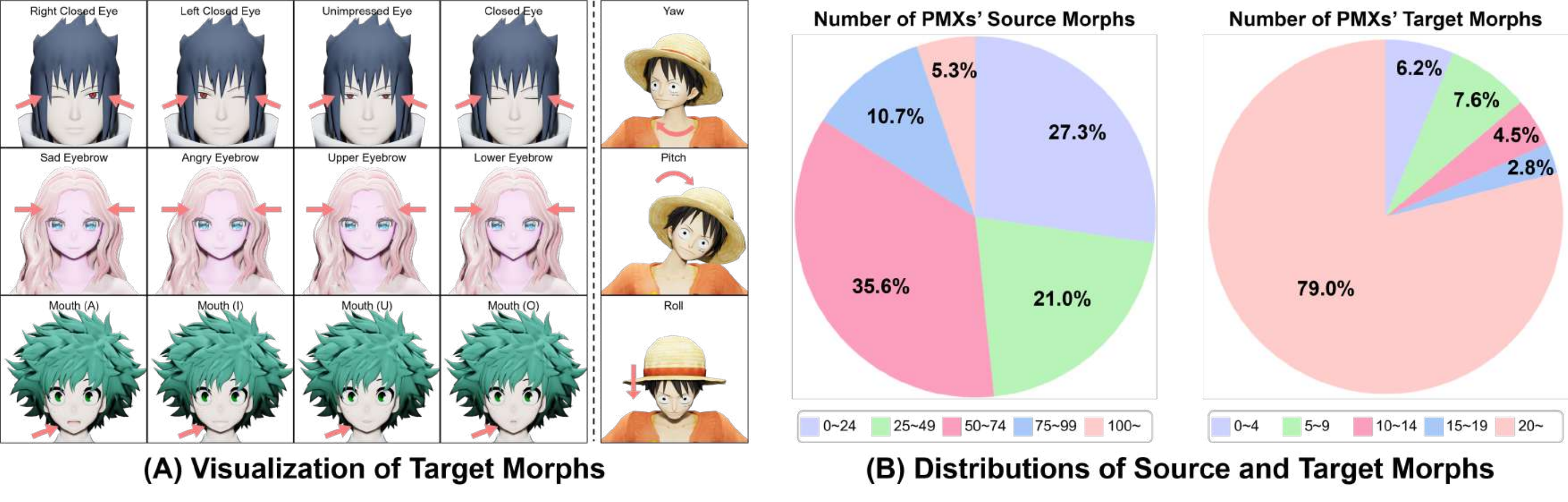}
    \vspace{-0.7cm}
    \caption{(A) Visualizing target morphs' examples and head rotation. 
    (B) The percentage of the number of source and target morphs on 3D animation models. 
    The number of source morphs are widely distributed ranging from 0 to over 100, and most animation models have dense usable annotations (\textit{i.e.}, target morphs).}
    \vspace{-0.75cm}
    \label{fig:morph}
\end{figure}

\noindent\textbf{Semantic Annotation (C).}
Each 3D animation model has a significantly different number of morphs ranging from zero to even over 100.
However, a morph naming convention is different according to a creator, which makes it difficult to apply a standardized criterion before annotating an accurate semantic of an individual morph.
A goal of the semantic annotation is to \textit{identify} expression-related morphs and \textit{annotate} the morphs according to the unified naming convention.
Importantly, this allows to sample a properly functioning expression-related source morph from a 3D animation model during rendering.
For example, when a morph \begin{CJK}{UTF8}{maru}あ\end{CJK} attached to a specific 3D animation model is identified as indicating a semantic of pronouncing the syllable \lq{ah}\rq\ with a mouth, then it can be annotated as the target morph (\textit{i.e.}, Mouth (A)).
After annotation, that source morph \begin{CJK}{UTF8}{maru}あ\end{CJK} of the 3D model is used, when the target morph Mouth (A) is determined to control the mouth shape. 

To achieve the semantic annotation, we first define 23 \textit{target} morphs, these are deemed as meaningful semantics to represent the facial expressions.
We select the target morphs out of candidates collaborated with animation experts who work with cartoon makers.
Fig.~\ref{fig:morph} (A) shows the examples of the target morphs that include meaningful semantics for three parts: eyes, eyebrows, and a mouth.
Conversely to the target morphs, we denote the original morphs as \textit{source} morphs in the remainder of this section.
Next, we attempt to match the source morphs to the target morphs.
Fortunately, a group of the source morphs with the identical name tends to portray the same semantics.
Therefore, we take a two-stage approach: a group annotation and an individual inspection. 
The former collectively match a group of the source morphs under the same name to a target morph; the latter is responsible for inspecting the matched source morphs one-by-one to confirm whether it works correctly.
During the group annotation, we count the number of source morphs that 3D models have, and remove the source morphs under 50.
The individual inspection reduces the erroneous annotations that occur at the group annotation.

For this, we first render the head images after applying the entire source morphs independently and a neutral image without applying any morph using a 3D animation model (Fig.~\ref{fig:pipeline} (C) upper part).
Afterwards, we match a group of source morphs to one of the target morphs (\textit{i.e.}, group annotation) and correct the results in a single morph-level via comparing a neutral and a morph-applied image for each source morph (\textit{i.e.}, individual inspection).
The entire procedure is conducted on the newly developed annotation system (Fig.~\ref{fig:pipeline} (C) lower part).
We provide the details of the defined target morphs and annotation system in the supplementary material.

% Dataset Table
\begin{table}[t!]
    \centering
    \scriptsize
    \begin{tabular}{c|cccccc}
        \toprule
         \textbf{Dataset} & \makecell[c]{\textbf{Num. of} \\ \textbf{Images}} & \makecell[c]{\textbf{Identity} \\ \textbf{Labels}} & \makecell[c]{\textbf{Face} \\ \textbf{Align.}} & \makecell[c]{\textbf{Unified} \\ \textbf{Style}}  & \makecell[c]{\textbf{Image} \\ \textbf{Source}} & \textbf{Attribute Anno.} \\
        \midrule
        \makecell[l]{Kaggle Anime Face~\cite{kaggleanimeface}} & 63K &  \xmark & \xmark & \xmark & Media & - \\
        \makecell[l]{Danbooru 2019~\cite{danbooru2019Figures}} & 302K & \cmark & \xmark & \xmark & Media & - \\
        \makecell[l]{iCartoonFace~\cite{zheng2020cartoon}} & 0.39M &  \cmark & \xmark & \xmark & Media & \makecell{3D Head Pose, \\ Bounding Box \\ Gender} \\
        \midrule
        \textbf{\makecell[l]{AnimeCeleb \\ (Ours)}} & 2.4M & \cmark & \cmark & \cmark & \makecell[c]{3D Models} & \makecell{3D Head Pose, \\ Expression, \\ Foreground Mask \\ Artistic Style} \\
        \bottomrule
    \end{tabular}
    \caption{Comparison between the AnimeCeleb and public animation head datasets.}
    \vspace{-0.9cm}
    \label{Table:dataset}
\end{table}

% Figure - iCartoonFace
\begin{figure}[t!]
    \centering
    \includegraphics[width=1.0\linewidth]{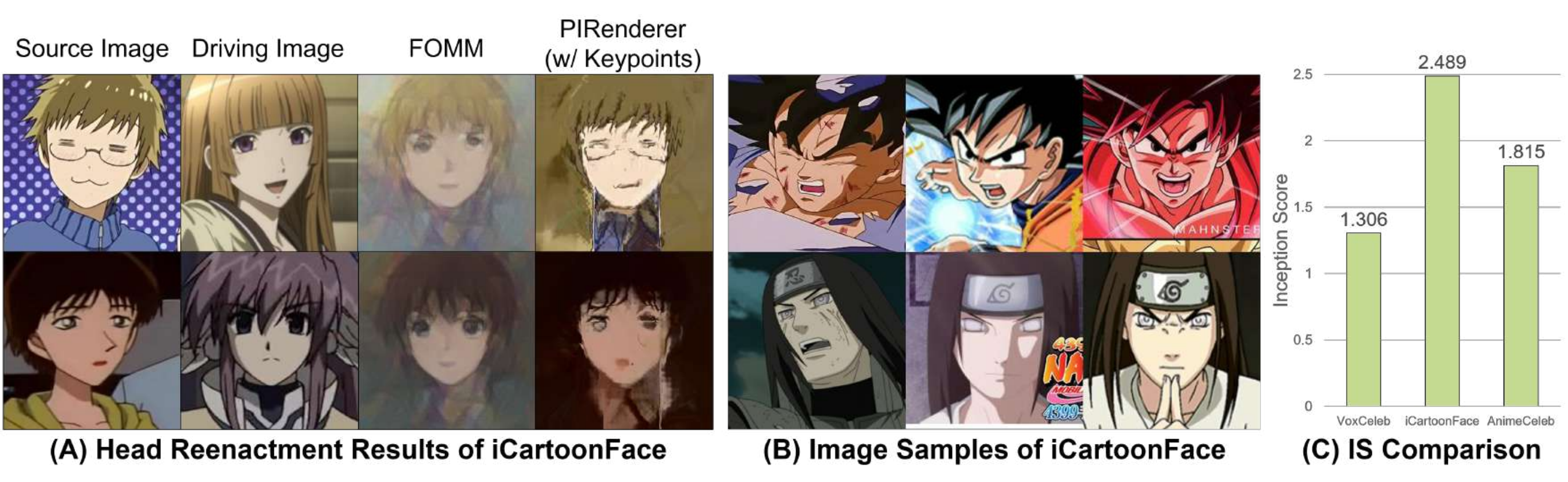}
    \vspace{-0.7cm}
    \caption{(A) Head reenactment results trained with the iCartoonFace that bear an identity leakage problem. (B) An intra-variation within the same identity of the iCartoonFace is extremely large. (C) Average inception score comparison on three datasets; the average scores using 1000 identities indicate that iCartoonFace contains relatively inconsistent styles within the identity than those of the VoxCeleb and the AnimeCeleb.}
    \vspace{-0.7cm}
    \label{fig:icartoonface}
\end{figure}

\noindent\textbf{Data Sampling (D).}
Throughout the data sampling, randomly selected target morphs for each part (\textit{i.e.,} eyes, eyebrows and a mouth) are applied to a 3D animation model.
The magnitudes of the morphs are determined by sampling from a uniform distribution, $\mathcal{U}(0,1)$, independently.
In respect to the head rotation, a 3D rotation matrix is computed taking yaw, pitch and roll values sampled between -20$^{\circ}$ and 20$^{\circ}$.
We render a transformed head after applying the morphs and the rotation, and also acquire a paired pose vector $\mathbf{p} \in \mathbb{R}^{20}$.
A detailed description of the pose sampling process is provided in supplementary material.

A real-time rendering engine that Blender provides is used to produce the manipulated images and paired pose vectors.
During rendering, we utilize 4 different types of shaders as shown in Fig.~\ref{fig:pipeline} to provide diverse textured 2D images.
Since the morphs and the head rotation are applied independently, two image groups: a group of frontalized images with expression (\textit{frontalized-expression}) and head rotated images with expression (\textit{rotated-expression}) are included in the AnimeCeleb.
The number of images sampled from the 3D model are determined differently depending on the number of annotated target morphs that a 3D animation model has.
When a 3D animation model contains more than five annotated target morphs, we generate 100 images; if not (\textit{e.g.}, zero), just 20 images are obtained.

\subsection{Dataset Description}\label{dataset_description}
\noindent\textbf{AnimeCeleb Properties.} 
Fig.~\ref{fig:morph} (A) shows the examples of multiple target morphs for each part and head rotation results.
The target morphs consist of 9 eye-related morphs, 9 eyebrow-related morphs and 5 mouth-related morphs.
Note that the pre-defined target morphs include the semantics related to both eyes or eyebrows, which fill two values (\textit{e.g.}, left and right eye) of a 17-dimensional pose vector (\textit{expression} part).
In total, 3613 different 3D models are used to generate the AnimeCeleb.
As can be seen in Fig.~\ref{fig:morph} (B) left, the number of source morphs of collected raw 3D animation models are widely distributed, averaging 49 morphs.
After the semantic annotation, most animation models have more than 20 target morphs as shown in Fig.~\ref{fig:morph} (B) right; this indicates the source morphs are densely matched to the target morphs.

\noindent\textbf{Comparison with Other Datasets.} 
As shown in Table~\ref{Table:dataset}, the AnimeCeleb has three advantages compared to the public existing animation head datasets~\cite{kaggleanimeface,danbooru2019Figures,zheng2020cartoon}.
The advantages mainly stem from exploiting the power of 3D software and 3D animation models.
First, detailed annotations such as facial expressions and head rotations can be easily gained because we are able to manipulate the head using our morph annotation (Table~\ref{Table:dataset} Attribute Anno.). 
Second, the AnimeCeleb provides a massive amount of animation images that have unified styles (Table~\ref{Table:dataset} Num. of Images, Unified Style).
We believe that these properties help to develop high-performing neural networks in broad applications.
Lastly, the AnimeCeleb contains four different unified styles in consideration of different cartoon textures.
A similar approach~\cite{Khungurn:2021} has been proposed using 3D animation models to construct an animation face dataset, and achieve a promising results on head reenactment.
The contribution of AnimeCeleb is the first publicly available dataset that contains animation faces with pose annotations as well as the data sampling pipeline.

\subsection{Animation Head Reenactment}\label{animation_head_reenactment}

\noindent\textbf{Overview.}
The head reenactment aims to transfer a pose from a driving image to a source image.
A common training scheme of the head reenactment model is to extract a pose from a driving image, and feed it with a source image to a decoder to reconstruct the driving image.
Therefore, training a high-performing head reenactment model requires a large-scale video dataset, containing a set of the same identity images that can serve as a source and driving image pair.
In a human domain, the VoxCeleb~\cite{nagrani2017voxceleb}, a large-scale talking head dataset, plays this role.
We believe that the AnimeCeleb is analogous to the VoxCeleb in an animation domain, which bears a potential to train a high-performing animation head reenactment model.

Prior head reenactment approaches are categorized into two groups whether a pre-computed pose annotation is utilized during training or not.
The FOMM does not use the pose annotation, and learn relative motion between two images to convey the pose to a source image.
In contrast, numerous studies~\cite{fewadvneural,ren2021pirenderer,fastbilayer} take advantage of the pose annotations such as keypoints and 3DMM parameters obtained from off-the-shelf pose extractors. 
Among them, we train two representative head reenactment baselines~\cite{firstorder,ren2021pirenderer} from each category with the AnimeCeleb: the FOMM~\cite{firstorder} and the PIRenderer~\cite{ren2021pirenderer}, which uses 3DMM parameters to describe a head pose.

% Figure - AnimeCeleb comparison & Control & Interpolation
\begin{figure}[t!]
    \centering
    \includegraphics[width=1.0\linewidth]{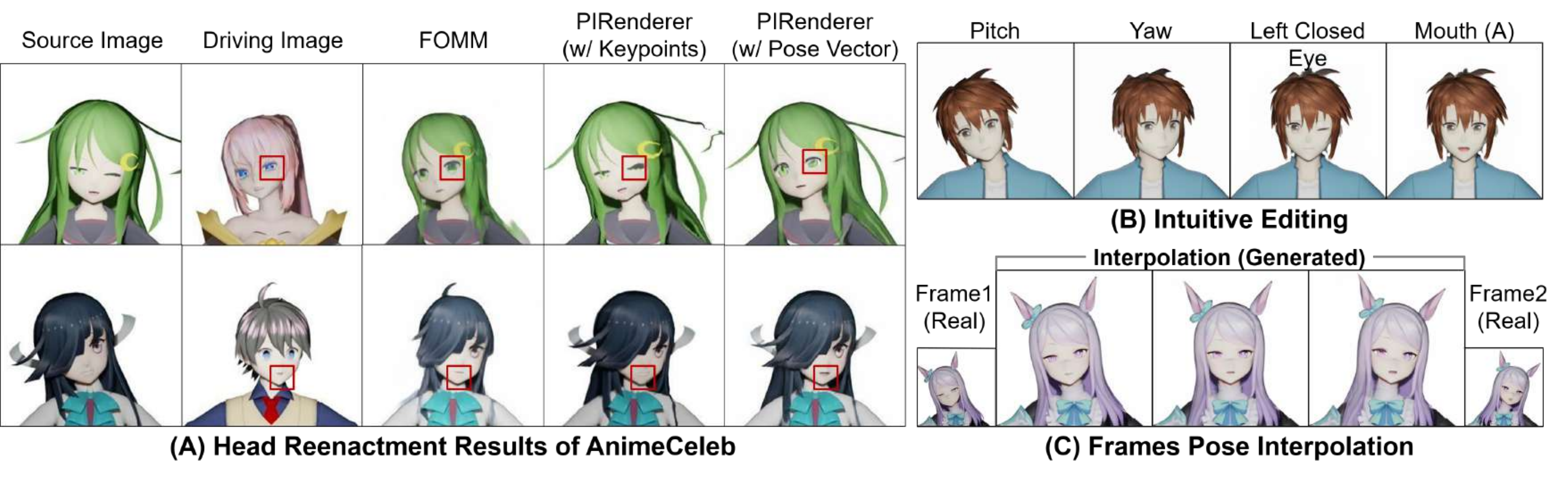}
    \vspace{-0.7cm}
    \caption{(A) Qualitative results of the FOMM and PIRenderer trained with the AnimeCeleb. (B) Intuitive editing of an animation head image with different pose vectors. (C) Filling in-between frames using linearly interpolated pose vectors.}
    \vspace{-0.3cm}
    \label{fig:baseline}
\end{figure}

% Table - AnimeCeleb Baselines
% Animation Face Reenactment Experiments
\begin{table}[t!]
    \small
    \centering
    \begin{tabular}{lccc}
    \toprule
    \multicolumn{1}{c}{\multirow{2}{*}{\textbf{Model}}} & \multicolumn{2}{c}{\textbf{Same-Identity}} 
    & \multicolumn{1}{c}{\textbf{Cross-Identity}} \\
    \cmidrule(lr){2-3} \cmidrule(lr){4-4}
    & FID$_{\downarrow}$ & SSIM$_{\uparrow}$ & FID$_{\downarrow}$\\
    \midrule
    FOMM & 23.45 &\ 0.824 &\ 29.94 \\
    PIRenderer(w/ keypoints) & 27.84 &\ 0.770 &\ 21.48 \\
    PIRenderer(w/ pose vector) & \textbf{20.27} &\ \textbf{0.826} &\ \textbf{16.52} \\
    \bottomrule
    \end{tabular}
    \caption{Quantitative results of animation head reenactment.
    Obviously, for the AnimeCeleb dataset, the PIRenderer trained with pose vector outperforms the PIRenderer with keypoints and the FOMM.}
    \vspace{-1.0cm}
    \label{Table:talkinghead}
\end{table}

\noindent\textbf{Experiment Setup.}
When training the PIRenderer, we replace 3DMM with the pose vectors of the AnimeCeleb.
For the dataset comparison, we additionally train the baselines~\cite{firstorder,ren2021pirenderer} using the iCartoonFace\cite{zheng2020cartoon}.
Although there exist other animation head datasets \cite{kaggleanimeface,danbooru2019Figures}, we select the iCartoonFace as a comparison dataset, acknowledging the size of it and accurate identity labels.
Furthermore, with the aim of pose annotation comparison, we train the PIRenderer leveraging the keypoints for both datasets.
We utilize an off-the-shelf animation keypoint  detector\footnote{https://github.com/hysts/anime-face-detector} that gives 28 keypoints of an animation head image.
 All implementations are conducted following the hyperparameters denoted the papers with 3319 train set and 294 test dataset created with the first shader style.

We evaluate the trained models on (1) Self-identity task where the same character provides the source and driving image, and (2) Cross-identity task where two frames  of different character sampled from the AnimeCeleb serve as the source and driving image.
For evaluation, Frechet Inception Distance (FID)~\cite{heusel2017fid} and  Structural Similarity (SSIM)~\cite{wang2004ssim} are adopted to measure the generated images quality.
Note that the AnimeCeleb is applicable to other existing head reenactment models~\cite{fastbilayer,marionette,fewadvneural} that need image keypoints, yet we implement two representative baselines here.

\noindent\textbf{Experimental Results with the iCartoonFace.}
Fig.~\ref{fig:icartoonface} (A) shows the cross-identity head reenactment outputs of two models trained with the iCartoonFace.
Despite the attempts to train the FOMM and PIRenderer with the iCartoonFace, we have found that the trained models show poor performance, producing blurry outputs.
We assume that excessive variation within a single identity is the main cause of the results.
In fact, considering that the iCartoonFace consists of the images collected from different appearance scenes, most images have own properties as seen in Fig.~\ref{fig:icartoonface} (B).
For quantitative analysis, we measure the Inception Score (IS)~\cite{salimans2016improved} by averaging 1,000 image sets of the same identity.
As seen in Fig.~\ref{fig:icartoonface} (C), we confirm that the iCartoonFace records higher IS score, compared to the VoxCeleb and the AnimeCeleb.
This indicates that the iCartoonFace contains unacceptable appearance complexity, hence learning from such images goes beyond the capacity of existing head reenactment models.

\noindent\textbf{Advantage of Pose Annotation.}
As seen in Fig.~\ref{fig:baseline} (A), the FOMM trained with the AnimeCeleb produces plausible outputs, yet still has undesirable deformation.
Different from it, the trained PIRenderers successfully preserve the source head structure while imitating a given driving image with both pose annotations (\textit{i.e.}, keypoints and pose vector).
Especially, the PIRenderer(w/ pose vector) accurately conveys a driving pose to the source image as shown in Fig.~\ref{fig:baseline} (A) red boxes. 
It is because the AnimeCeleb pose vectors hold more direct guidance (\textit{e.g.}, 80\% mouth openness) than the keypoints.
This results can be quantitatively confirmed in Table~\ref{Table:talkinghead}, where the PIRenderer(w/ pose vector) outperforms other baselines on both same-identity and cross-identity head reenactment tasks.
Besides, the PIRenderer(w/ pose vector) is able to intuitively edit head poses based on given pose vectors (Fig.~\ref{fig:baseline} (B)) and generate the in-between frames by interpolating the pose vectors of two different frames (Fig.~\ref{fig:baseline} (C)).

% Figure - Pose Distribution Visualization
\begin{figure}[t!]
    \centering
    \includegraphics[width=1.0\linewidth]{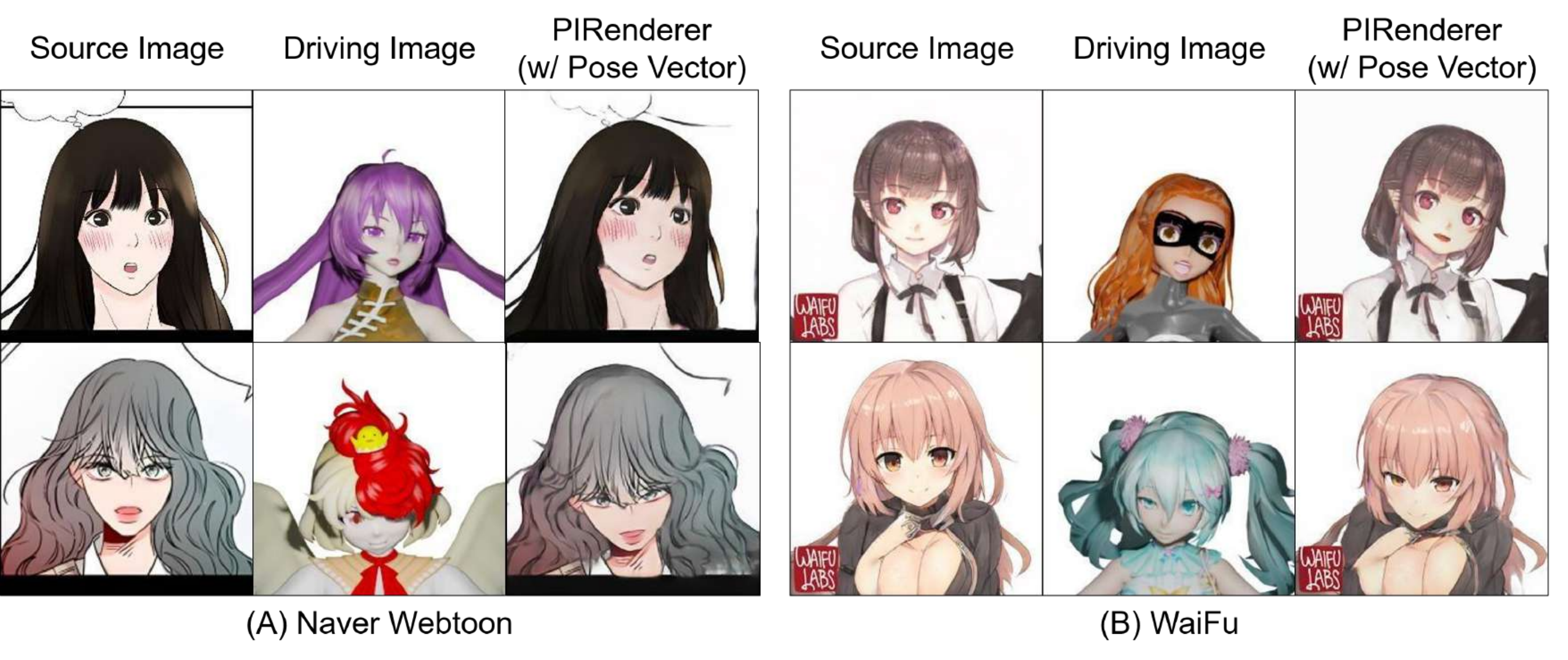}
    \vspace{-0.7cm}
    \caption{Head reenactment results on other animation datasets.}
    \vspace{-0.8cm}
    \label{fig:out-domain}
\end{figure}

\noindent\textbf{Other Animation Results.}
We demonstrate the generalization capacity of the trained model on other animation datasets.
In an experiment, we evaluate the PIRenderer(w/ pose vector) on different collected head datasets including Waifu Labs~\footnote{https://waifulabs.com/} and Naver Webtoons~\footnote{https://comic.naver.com/}.
As seen in Fig.~\ref{fig:out-domain}, the model successfully transfer a given driving image pose to an animation head.
We provide the details of the collected animation head datasets and additional results on other examples in supplementary material.

\vspace{-0.3cm}
\section{Cross-Domain Head Reenactment}\label{cdhr}
\vspace{-0.1cm}
\noindent\textbf{Overview.}
Although we show a promising animation head reenactment result in Section \ref{animation_head_reenactment}, controlling characters' head pose as a human user wants (\textit{i.e.}, cross-domain head reenactment) is another important application that bears a potential to be used in a virtual YouTuber system and a cartoon production.
In this section, we address the cross-domain head reenactment using the proposed pose mapping method and the \model.

In a standard head reenactment training scheme, two frames are sampled from a video: a source image $s$ and driving image $d$, and reconstruct $d$. 
Different from previous methods~\cite{firstorder,ren2021pirenderer}, we leverage two videos from different domains, respectively.
Since a direct supervision across domains is not available during training, the source and driving image pair from animation domain: $s^{(a)}$, $d^{(a)}$ and human domain: $s^{(r)}$, $d^{(r)}$ are utilized to reconstruct the driving images, $d^{(a)}$ and $d^{(r)}$, respectively.
In the following, we illustrate the details of a driving pose representation(Section~\ref{subsec:driving_motion_descriptor}).
Then, we describe a training pipeline and its objective functions (Section~\ref{subsec:training_pipeline}).

\noindent\textbf{Difference from PIRenderer.}
Our architecture design is inspired by PIRenderer~\cite{ren2021pirenderer}, yet two novel components, a pose-mapping method and separate domain-specific networks, are proposed to improve cross-domain head reenactment performance. 
The pose-mapping method enables to align blendshape and 3DMM, which gives the capability to handle a pose from human domain (\textit{i.e.}, cross domain). 
Also, the domain-specific networks help to preserve a given source image’s textures for each domain, and improve the quality of image. 
Note that our pose-mapping method can help PIRenderer to improve the performance on cross-domain head reenactment task.

\subsection{Driving Pose Representations}\label{subsec:driving_motion_descriptor}

\noindent\textbf{Human Pose Representation.}
Our approach employs the 3DMM parameters to describe a pose of a driving human head image.
With the 3DMM, a 3D human face shape $\mathbf{S}$ can be represented as $\mathbf{S} = \bar{\mathbf{S}} + \alpha \mathbf{B}_{id} + \beta \mathbf{B}_{exp}$,
where $\bar{\mathbf{S}}$ is the average face shape, $\mathbf{B}_{id}$ and $\mathbf{B}_{exp}$ denote the principal components of identity and expression based on 200 scans of human faces~\cite{blanz1999morphable}, respectively.
Also, $\alpha \in \mathbb{R}^{80}$ and $\beta \in \mathbb{R}^{64}$ indicate the coefficients that control the relative magnitude between the facial shape and expression basis.
The head rotation and translation are defined as $\mathbf{R} \in SO(3)$ and $\mathbf{t} \in \mathbb{R}^{3}$.
We use a pre-trained 3D face reconstruction model~\cite{deng2019accurate} to extract the 3DMM parameters from the human head images.
Discarding $\alpha$ for excluding an identity-related information, we only exploit a subset space of the 3DMM parameters $\mathcal{M}$ to represent a human head pose, where $\mathbf{m} \in \mathcal{M}$ comprises of expression coefficients, head rotation and translation: $\mathbf{m} \equiv \{\beta, \mathbf{R}, \mathbf{t}\} \in \mathbb{R}^{70}$.

\begin{figure}[t!]
    \centering
    \includegraphics[width=1.0\linewidth]{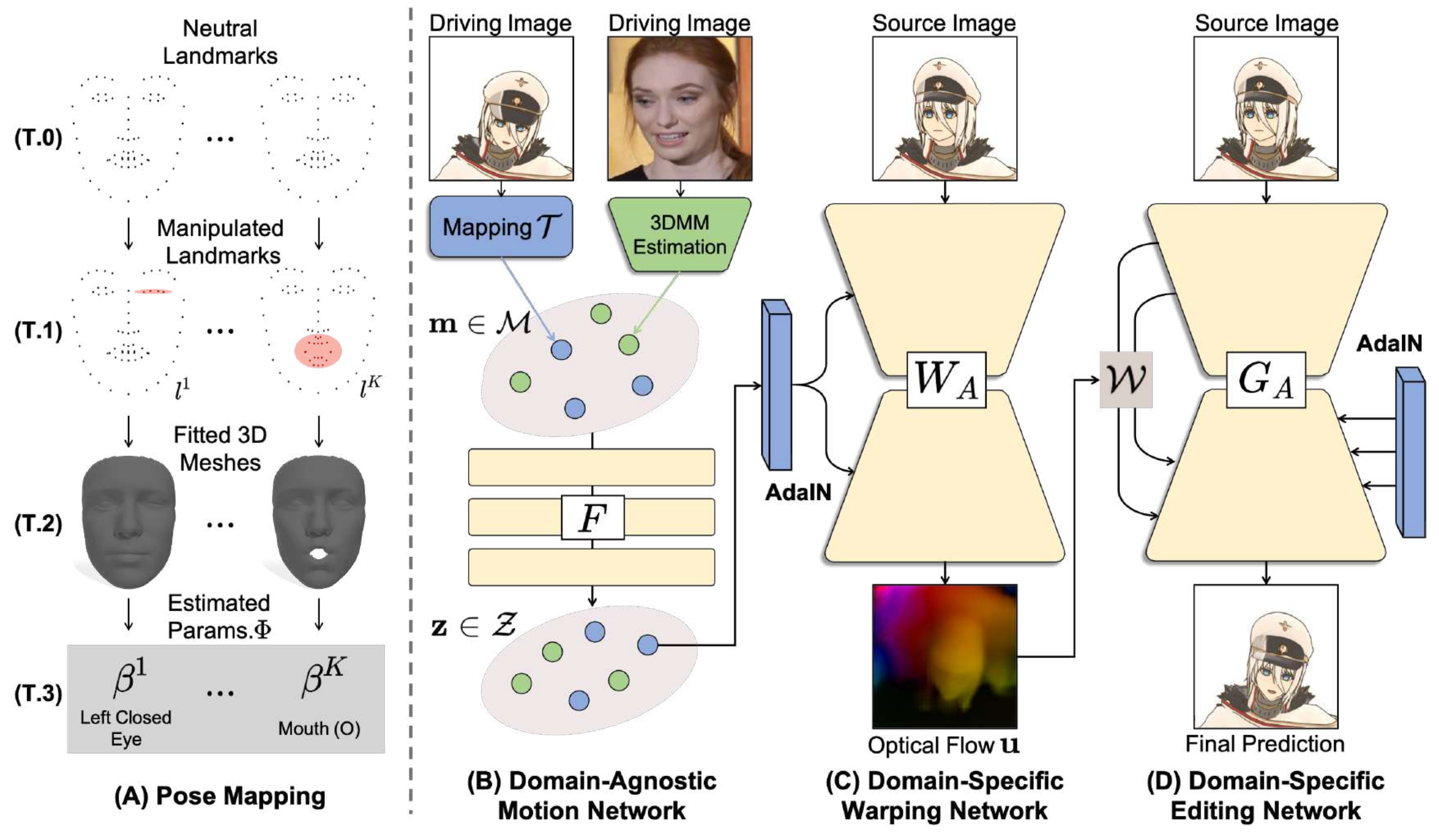}
    \vspace{-0.75cm}
    \caption{Overview of (A) pose mapping method and (B)-(D) \model.}
    \vspace{-0.9cm}
    \label{fig:model}
\end{figure}

\noindent\textbf{Pose Mapping.} 
The AnimeCeleb pose vector $\mathbf{p} \in \mathbb{R}^{20}$ consists of independent coefficients $\mathbf{b} \in \mathcal{B}$ and head angles $\mathbf{h} \in \mathcal{H}$, where $\mathcal{B}$ denotes a 17-dimensional space of concatenated expression coefficient and $\mathcal{H}$ indicates a 3D head angle space.
In this step, we aim at discovering a mapping relationship from the AnimeCeleb pose vector to the 3DMM parameters.
To this end, we propose a pose mapping function: $\mathcal{T}: \mathcal{B} \times \mathcal{H} \rightarrow \mathcal{M}$, which is responsible to find its corresponding 3DMM parameters, given a pose vector.
We construct a direct mapping relationship between the coefficients $\mathbf{b}$ and the 3DMM expression parameters $\beta$ using facial landmarks as a proxy space and expressing the each coefficient's semantics via manually manipulating the landmark positions.
In the following, we elaborate the details step-by-step with Fig.~\ref{fig:model} (A).

(T.0) Before the landmark manipulation, we first obtain an initial landmark position, which corresponds to a neutral 3DMM coefficient.
To be specific, the initial landmark position is obtained from a rendered mesh with setting the entire 3DMM coefficients as $\mathbf{0}$ expressed as $\{\alpha_0,\beta_0,\mathbf{R}_0,\mathbf{t}_0\}$, meaning that the average face shape $\bar{\mathbf{S}}$ at center location offers the initial landmark position.
(T.1) Next, the initial landmarks are manipulated according to each  semantic; for example, \textit{left closed eye} landmarks can be achieved by minimizing the distances between the upper and the lower eyelid keypoints at the left eye.
(T.2 and T.3) Then, the manipulated landmarks $l^k$ with $k$-th semantic are used to update the initial $\beta$ by minimizing the $\ell_2$ distance between $l^k$ and the landmarks extracted from the rendered mesh using $\beta$.
Also, we employ a $\ell_2$ regularization during updating $\beta$.
Completing this process for each landmark, we can gain the fitted 3DMM expression parameters for each semantic: $\Phi = \{\beta^{k}\}_{k=1}^{17} \in \mathbb{R}^{17 \times 64}$.
Finally, the pose mapping function can be written as:
$\mathbf{m}_{i} = \mathcal{T}(\mathbf{b}_{i}, \mathbf{h}_{i}) = (\mathbf{b}_{i}\cdot\Phi) \oplus \Pi(\mathbf{h}_{i}) \oplus \mathbf{0} \in \mathcal{M},$
% \begin{equation}
%     \mathbf{m}_{i} = \
%         \mathcal{T}(\mathbf{b}_{i}, \mathbf{h}_{i}) = \
%         (\mathbf{b}_{i}\cdot\Phi) \oplus \Pi(\mathbf{h}_{i}) \oplus \mathbf{0}
%         \in \mathcal{M},
% \end{equation}
where $\Pi$ denotes a mapping to convert a degree into radian measurement and $\oplus, i$ indicate a concatenation operation and a data index, respectively.
In addition, $\mathbf{0} \in \mathbb{R}^{3}$ is concatenated to represent translation parameters.

% Figure - Overview of our framework

\subsection{Training Pipeline}\label{subsec:training_pipeline}
Fig.~\ref{fig:model} depicts an overview of our framework, which consists of three networks described below. 

\noindent\textbf{Motion Network.} 
Given a driving pose $\mathbf{m}$, our motion network $F$ generates a latent pose code $\mathbf{z} \in \mathcal{Z}$, where $\mathcal{Z}$ denotes a latent pose space.
Formally, this can be written as: 
$ \mathbf{z}^{(a)} = F(\mathbf{m}^{(a)}), \mathbf{z}^{(r)} = F(\mathbf{m}^{(r)}),$
where $\mathbf{m}^{(a)} = \mathcal{T}(\mathbf{b},\mathbf{h})$ is a transformed driving pose corresponding to the driving image $d^{(a)}$ in an animation domain and $\mathbf{m}^{(r)}$ denotes a subset of 3DMM paramters obtained from the driving image $d^{(r)}$ in a human domain, respectively.
Thanks to the pose mapping method, the motion network $F$ can be designed as \textit{domain-agnostic} manner.
The learned latent pose code $\mathbf{z}$ is transformed to estimate the affine parameters for adaptive instance normalization (AdaIN)~\cite{huang2017arbitrary} operations. 
The pose information parameterized as the affine parameters plays a role in predicting an optical flow in the warping network $W$ and injecting a fine-detailed pose in the editing network $G$.

\noindent\textbf{Warping \& Editing Network.}
For sake of simplicity, we omit the domain notation unless needed, such as $\mathbf{z}=\{\mathbf{z}^{(a)}, \mathbf{z}^{(r)}\}$, $d=\{d^{(a)},d^{(r)}\}$, and $s=\{s^{(a)},s^{(r)}\}$ in the descriptions of warping and editing network.
Inspired by the PIRenderer~\cite{ren2021pirenderer}, we employ \textit{domain-specific} warping networks and an editing network for each domain.
A warping network $W$ takes a source image $s$ and latent pose code $\mathbf{z}$ to predict the optical flow $\mathbf{u}$ that approximates the coordinate offsets to reposition a source head alike a driving head.

Next, the source image is fed into an encoder part of a editing network $G$ and the optical flow $\mathbf{u}$ is applied to the intermediate multi-scale feature maps.
This leads to spatial deformation of the feature maps according to the driving pose.
During decoding in $G$, the AdaIN operation is used to inject the pose information.
After training, the warping network mainly focuses on causing a large pose, including the head rotation, whereas the editing network serves to portrait an detailed expression-related pose.
We train our framework with a reconstruction loss and a style loss following the PIRenderer~\cite{ren2021pirenderer}.
The architecture, implementation details and objective functions are elaborated in the supplementary material.

% Table - Cross-Domain Baselines
% Comparison with Baselines
\begin{table*}[t!]
    \scriptsize
    \centering
    \begin{tabular}{@{} clccccccc @{}}
    \toprule
    \multirow{2}{*}{\textbf{Train Dataset}} & 
    \multicolumn{1}{c}{\multirow{2}{*}{\textbf{Methods}}} & 
    \multicolumn{2}{c}{\shortstack{\textbf{Self-identity}  \\ \textbf{(AnimeCeleb)}}} & 
    \multicolumn{2}{c}{\shortstack{\textbf{Self-identity}  \\ \textbf{(VoxCeleb)}}} & 
    \multicolumn{2}{c}{\shortstack{\textbf{Cross-domain}  \\ \textbf{(Vox.}$\rightarrow$\textbf{Anime.})}} & 
    \multicolumn{1}{c}{\shortstack{\textbf{Cross-domain}  \\ \textbf{(Anime.}$\rightarrow$\textbf{Vox.})}} \\
    \cmidrule(lr){3-4} \cmidrule(lr){5-6} \cmidrule(lr){7-8} \cmidrule(lr){9-9}
     && FID$_{\downarrow}$ & SSIM$_{\uparrow}$ 
     & FID$_{\downarrow}$ & SSIM$_{\uparrow}$
     & FID$_{\downarrow}$ & HAE$_{\downarrow}$ & FID$_{\downarrow}$ \\
    \midrule
    \multirowcell{3}{\textit{Single Dataset} \\ \textit{(VoxCeleb)}}
    & FOMM
     & 47.91 &\ 0.648 
     & \textbf{16.10} &\ \textbf{0.803} 
     & 122.83 &\ 0.177 &\ 94.23 \\
    & PIRenderer
     & 134.91 &\ 0.532 
     & 19.67 &\ 0.604
     & 95.75 &\ 0.176 &\ 96.42 \\
    & LPD
     & - &\ - 
     & - &\ -  
     & 166.54 &\ 0.171 &\ - \\
    \midrule
    \multirowcell{4}{\textit{Joint Datasets} \\ (\textit{AnimeCeleb},\\ \textit{VoxCeleb})}
    & FOMM
     & 45.01 &\ \textbf{0.748}
     & 19.60 &\ 0.748
     & 144.88 &\ 0.196 &\ 126.49 \\
    & PIRenderer+$\mathcal{T}$
     & 16.07 &\ 0.735
     & 18.98 &\ 0.611
     & 69.80 &\ 0.195 &\ 61.67 \\
    \cmidrule(lr){2-9}
    & \textbf{Ours}
     & \textbf{16.05} &\ 0.738
     & 19.34 &\ 0.606
     & \textbf{18.78} &\ \textbf{0.128} &\ \textbf{41.04} \\
    \bottomrule
    \end{tabular}
    \caption{Quantitative comparison with baselines on self-identity and cross-domain head reenactment tasks. The expression A $\rightarrow$ B denotes that transferring a A's motion to B's a source image.}
    \vspace{-0.8cm}
    \label{Table:cross_baseline}
\end{table*}

\vspace{-1.0em}
\subsection{Experiments}

\noindent\textbf{Experiment Setup.}
Different from Section~\ref{animation_head_reenactment}, we use both cartoon texture shader style AnimeCeleb and the VoxCeleb\cite{nagrani2017voxceleb} as a training dataset.
The VoxCeleb contains 22,496 talking-head videos collected from online videos, and we use downloadable 18,503 videos for the train set and 504 videos for test set.

We evaluate the trained models on self-identity, and cross-domain head reenactment where the images of the AnimeCeleb and the VoxCeleb alternatively serve as a source and a driving image respectively.
Similar to Section~\ref{animation_head_reenactment}, FID and SSIM are used to assess the quality of generated images.
In addition, we introduce a Head Angle Error (HAE) that measures the $\ell_{1}$ distances between the driving image's head angles and those of the generated image with the aim of evaluating head rotating ability.
To be specific, we take advantage of a pre-trained head angle regressor, based on ResNet-18~\cite{he2016deep} architecture and trained with the AnimeCeleb train set using $\ell_{1}$ distance objective function between a predicted angle and the ground-truth $\mathbf{h}$.
In experiments, we use randomly sampled 1,000 pairs of source and driving images to compute evaluation metrics.

\noindent\textbf{Comparison with State of the Art.}\label{subsec:baselines}
We compare the the \model with state-of-the-art models~\cite{latentpose,firstorder,ren2021pirenderer} quantitatively and qualitatively.
Since we leverage two datasets during training, comparable baselines are trained on either the VoxCeleb following their original implementations or both the VoxCeleb and AnimeCeleb.
During evaluation, we make an inference of manipulated animation or human head by optionally leveraging the decoder of each domain.
We describe the details of the baselines and settings in the supplementary material.

\begin{figure}[t!]
    \centering
    \includegraphics[width=1.0\linewidth]{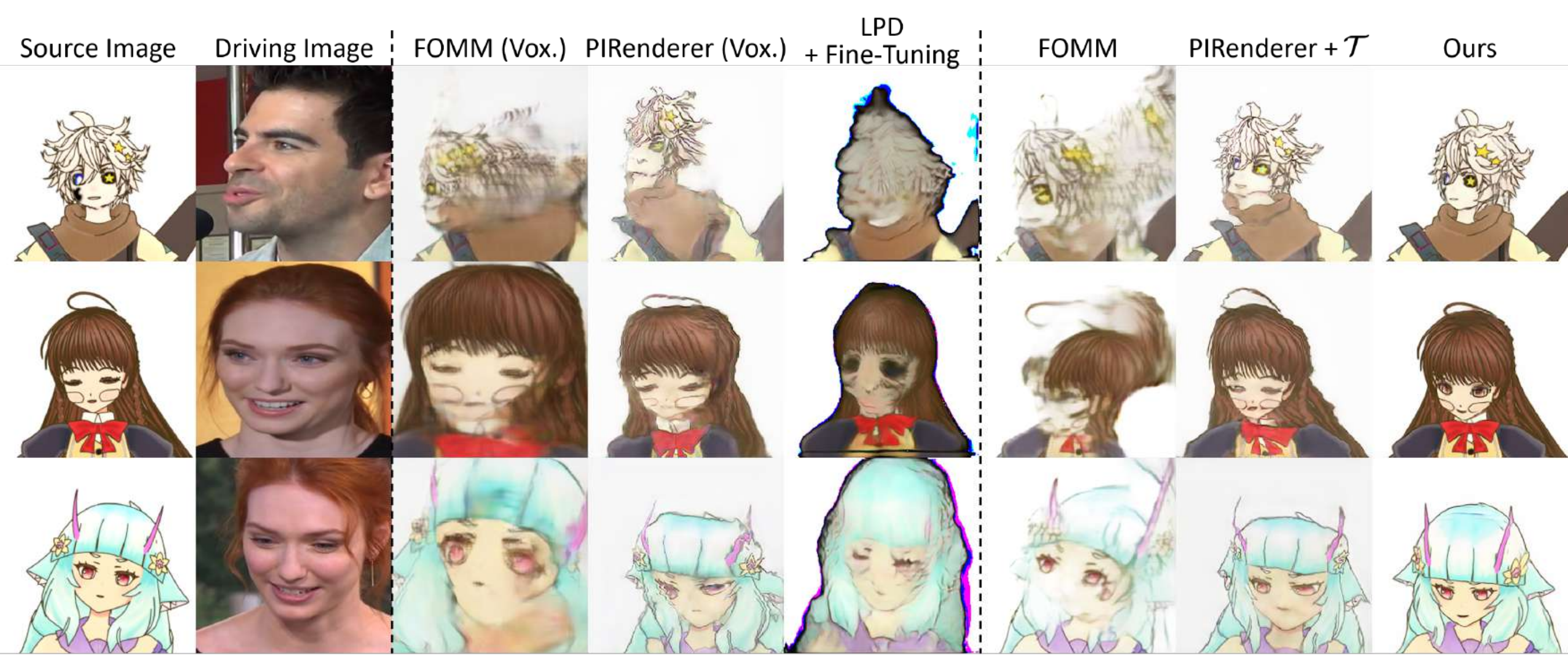}
    \vspace{-0.6cm}
    \caption{Qualitative comparison between our model and the baselines.}
    \vspace{-0.7cm}
    \label{fig:cross_baseline}
\end{figure}

Table~\ref{Table:cross_baseline} shows quantitative comparisons between the \model and the baselines on the self-identity and the cross-domain head reenactment.
When evaluating the self-identity head reenactment within the AnimeCeleb, it is obvious that the models trained on both the AnimeCeleb and the VoxCeleb surpass those trained on the VoxCeleb.
On the contrary, quantitative results on self-identity head reenactment within the VoxCeleb demonstrate that joint datasets may be harmful to the reconstruction task.
Unlike these results, our model outperforms all baselines on cross-domain head reenactment tasks in terms of an image quality and an imitating head pose, indicating the superiority of our model in transferring a pose across the domains.

Fig.~\ref{fig:cross_baseline} shows qualitative comparisons between the \model and the baselines on the cross-domain head reenactment. 
The FOMM, which relies on the unsupervised landmarks, does not work well, because the model attempts to align the appearance of the source image as the driving image's head structure, and this leads to the identity leakage problem as well as introducing blurring artifacts.
In contrast, the PIRenderer and latent pose descriptor (LPD)~\cite{latentpose}, where the pose is injected by the AdaIN operations, successfully retain a head structure of the source image, yet produce rather blurry outputs. 
As seen in the PIRenderer+$\mathcal{T}$, 
the blurry artifacts can be improved by incorporating the AnimeCeleb as an additional training dataset with the pose mapping $\mathcal{T}$.
Meanwhile, our model clearly outperforms the baselines, preserving more vivid textures of the source image and accurately reflecting the pose of the driving image with the aid of the domain-specific networks.
We conclude that the shared pose space introduced by the pose mapping and the domain-specific design help the model to transfer the pose across domains.
We include more results in the supplementary material.
\vspace{-0.3cm}
\section{Conclusions}~\label{conclusion}
In this paper, we present the AnimeCeleb, a large-scale animation head dataset, which is a valuable and practical resource for developing animation head reenactment model.
Departing from existing animation datasets, we utilize 3D animation models to construct our animation head dataset by simulating facial expressions and head rotation, resulting in neatly-organized animation head dataset with rich annotations.
For this purpose, we built a semi-automatic data creation pipeline based on Blender and a semantics annotation tool.
We believe that the AnimeCeleb would boost and contribute to animation-related research.
On the other hand, we propose the pose mapping and architecture to address cross-domain head reenactment to admit transferring a given human head motion to an animation head. 
Conducted experiments demonstrate the effectiveness of the \model on cross-domain head reenactment and intuitive image editing.
In the future work, we plan to extend the AnimeCeleb and develop more advanced cross-domain head reenactment model.

\noindent\textbf{Acknowledgements.}
This work was supported by the Institute of Information \& communications Technology Planning \& Evaluation (IITP) grant funded by the Korean government(MSIT) (No. 2019-0-00075, Artificial Intelligence Graduate School Program(KAIST), No. 2021-0-01778, Development of human image synthesis and discrimination technology below the perceptual threshold), and the Air Force Research Laboratory, under agreement number FA2386-22-1-4024. 
The U.S. Government is authorized to reproduce and distribute reprints for Governmental purposes notwithstanding any copyright notation thereon. 
Finally, we thank all researchers at NAVER WEBTOON Corp.

\clearpage

\bibliographystyle{splncs04}
\bibliography{reference}

\newpage
\appendix
\begin{center}
    \vspace*{1.1cm}
    \Large{\bf{Supplementary Material}}
    \vspace*{1.7cm}
\end{center}

\noindent\textbf{Overview of Supplementary Material.}~\label{overview}
This supplementary material consists of 6 sections: 
(1) related literature and context of our paper (Section~\ref{related_work}),
(2) details of data creation process (Section~\ref{animeceleb_details}),
(3) additional AnimeCeleb samples and experimental results (Section~\ref{animeceleb_additional}),
(4) implementation details of the \model and the baselines (Section~\ref{animo_details}),
(5) additional head reenactment results of \model (Section~\ref{animo_additional}), and 
(6) discussions and future work (Section~\ref{discussions}).

% =========================================================================== %

\section{Related Work}~\label{related_work}

With abundance of digital contents, numerous animation datasets collected from different media are released to community. 
Focusing on animation head datasets, there exist multiple studies~\cite{kaggleanimeface,danbooru2019Figures,fujimoto2016manga109,rios2021daf,zheng2020cartoon} that provide the pre-processed animation heads.
Based on these datasets, early animation-related research~\cite{qin2017faster,takayama2012face,zhang2020acfd} mainly focused on recognizing and detecting an animation character in animation scenes. 
However, an extension of animation research to generative modeling is non-trivial.
One major bottleneck is that the released datasets are collected from unlisted online source, thereby containing unexpected and noisy images (\textit{e.g.}, an occluded head).
In this regard, existing datasets are forced to narrow their application scope; for example, current animation datasets are not suitable to train \textit{head reenactment models}~\cite{latentpose,marionette,firstorder,onefreeview,fastbilayer,fewadvneural}.

Head reenactment aims to drive a source image to mimic a motion of a target image while preserving identity of the source image.
Most approaches~\cite{latentpose,marionette,firstorder,onefreeview,fastbilayer,fewadvneural} use two frames from the same video during training; an image conveys the identity-related information while the other provides the motion-related information, which are combined to produce a final output.
Also, multiple pose representations (\textit{e.g.}, keypoints and 3DMM parameters) play vital roles to deliver the head motion in previous literature~\cite{marionette,ren2021pirenderer,fastbilayer,fewadvneural}.
In fact, the pose representation is an important aspect for head reenactment approaches as shown in previous work~\cite{latentpose} when training a high-performing model. 

Undoubtedly, the collected animation images of current datasets~\cite{kaggleanimeface,danbooru2019Figures,fujimoto2016manga109,rios2021daf,zheng2020cartoon} do not include its detailed pose annotations, and obtaining accurate pose representations is also non-trivial. 
In opposition, AnimeCeleb provides numerous groups of images that have the same identity, and detailed pose annotations, which bear a potential to be used for various generation tasks.
\clearpage
\vspace{-0.5cm}
\section{Details of Data Creation Process}~\label{animeceleb_details}
\vspace{-0.5cm}

\begin{figure}[t!]
    \centering
    \includegraphics[width=1.0\linewidth]{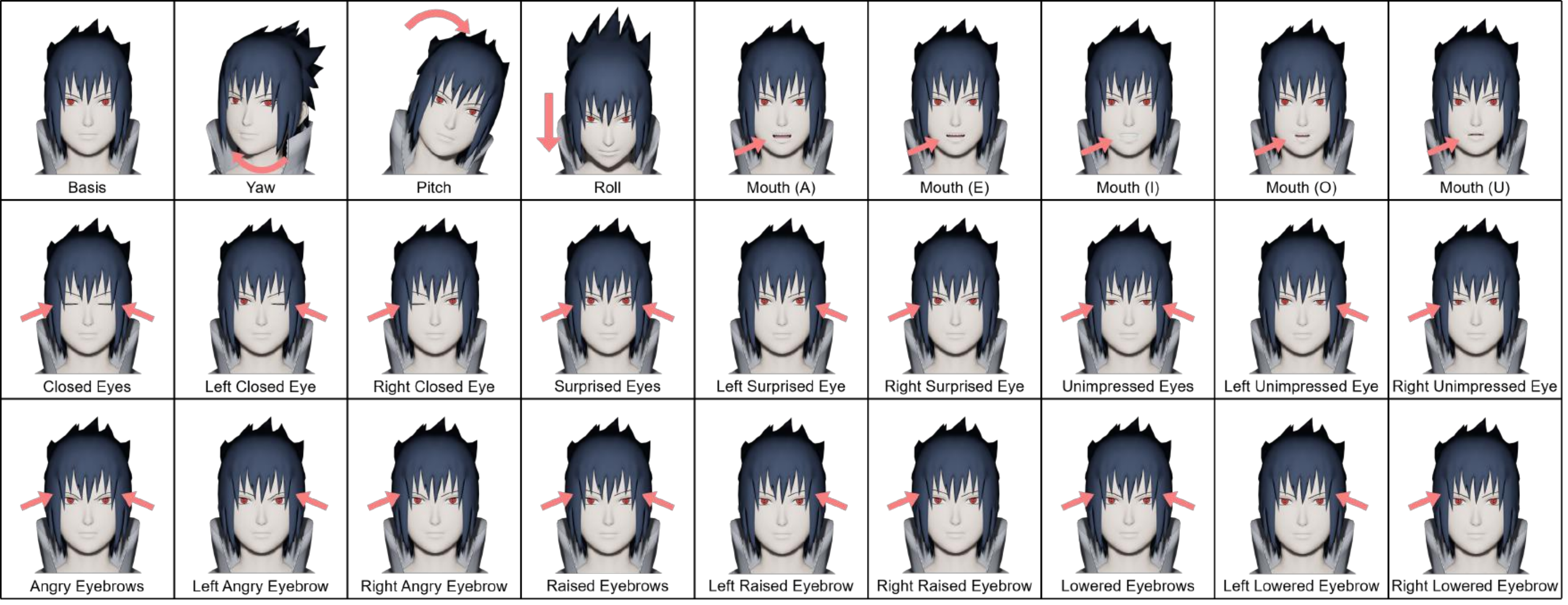}
    \vspace{-0.7cm}
    \caption{Visualizations of all target morphs and 3D head angles.
        Given a neutral image (\textit{Top-left}), we apply every annotated target morph independently with the maximum intensity (\textit{i.e.}, 1.0) to obtain the morph-applied images.
        We highlight the locations, where manipulations occur with \textcolor{pink}{pink} arrows.}
    \vspace{-0.7cm}
    \label{sup-fig:all-target-morphs}
\end{figure}

\begin{figure}[h!]
    \centering
    \includegraphics[width=0.7\linewidth]{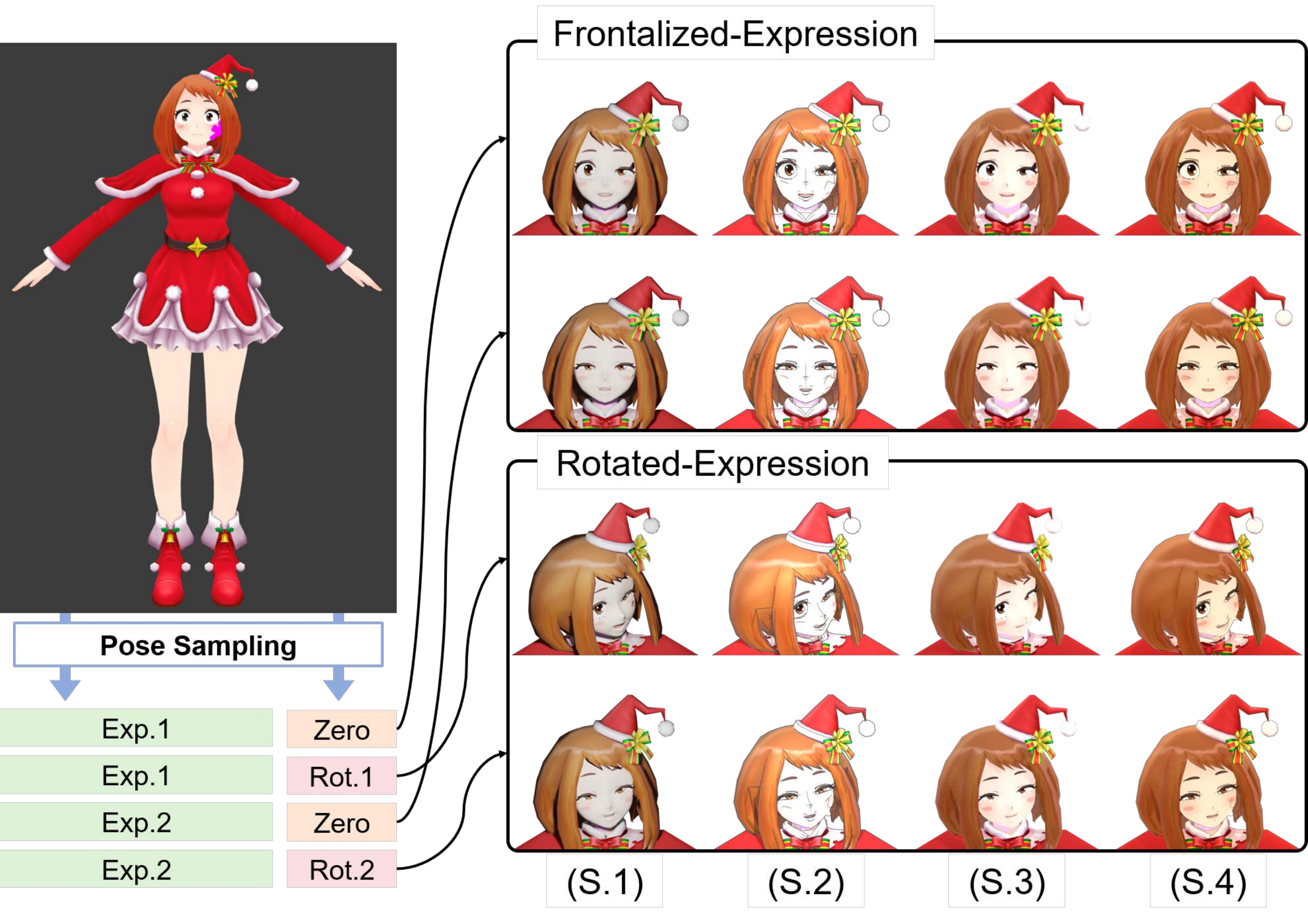}
    \vspace{-0.5cm}
    \caption{Examples of sampled data. 
    Given a 3D animation model, two groups of images are generated: (1) \textit{frontalized-expression} images using the sampled target morphs and zero head angles (\textit{Top-right}), and (2) \textit{rotated-expression} images after adding the sampled head angles (\textit{Bottom-right}).
    Note that four different shading styles are applied for image rendering.}
    \vspace{-0.5cm}
    \label{sup-fig:sampling}
\end{figure}

\begin{figure}[t!]
    \centering
    \includegraphics[width=0.675\linewidth]{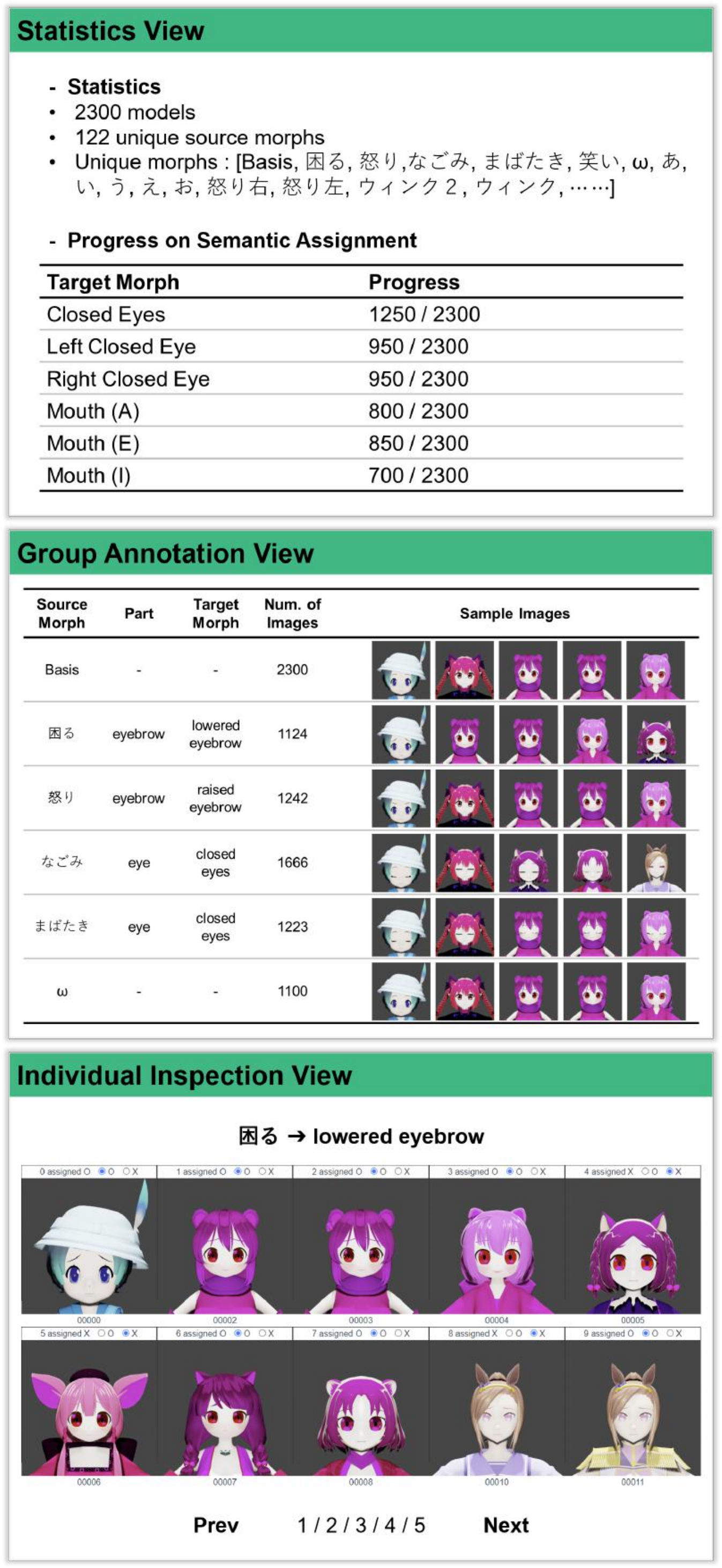}
    \vspace{-0.2cm}
    \caption{Simplified semantic annotation system overview.}
    \vspace{-0.3cm}
    \label{sup-fig:annotation_tool}
\end{figure}

In this section, we present the details of the data creation process as follows:
\begin{itemize}
    \item The visualizations of entire pre-defined target morphs that a single character has (Fig.~\ref{sup-fig:all-target-morphs}).
    \item Detailed user interfaces of the annotation system: statistics, group annotation, and individual inspection (Fig.~\ref{sup-fig:annotation_tool}) and mapping relationships between the source morphs and the target morphs after the annotation (Table~\ref{sup-table:mapping}).
    \item Detailed description of pose sampling process for generating a pose vector (Algo.~\ref{alg:pose_sampling}).
    \item Sampling examples from a 3D animation model (Fig.~\ref{sup-fig:sampling}).
\end{itemize}

\noindent\textbf{Visualizations of Target Morphs.}~\label{subsec:target_morphs}
Fig.~\ref{sup-fig:all-target-morphs} shows the visualizations of the manipulated poses and their corresponding target morphs, which are responsible for annotating the source morphs.
For head rotation and mouth annotation, there is a single value to control each semantic, respectively.
On the other hand, for eyes and eyebrows annotation, we consider left and right part separately and define three different target morphs: left-related, right-related and both-related semantics.
Note that although we have defined 23 target morphs including six morphs that control both parts (\textit{e.g.}, closed eyes and raised eyebrows), during constructing a pose vector, the both-related morphs simultaneously determine two values of the pose vector (\textit{i.e.}, left and right part).
Therefore, the dimensions of a pose vector become 20 (=17+3) with three additional head angle dimensions.

% Figure - Data Morphs

\noindent\textbf{Semantic Annotation System.}~\label{subsec:annotation_system}
Fig.~\ref{sup-fig:annotation_tool} shows the components of semantic annotation system developed with Vue.js~\footnote{https://vuejs.org/}.
Given a group of neutral images and morph-applied images, our system aims at visualizing the images and the source morph names.
Through the annotation, the source morphs are annotated as the target morphs, considering a semantic match.

The system consists of three views: statistics, group annotation and individual inspection.
In statistics view, there are the number of models and unique morph names that the models contain, and annotation progress shows the ratio of the annotated models to the total models.
During annotation, we match the source morphs (\textit{e.g.}, 困る and なごみ) as their corresponding target morphs (\textit{e.g.}, lowered eyebrows and closed eyes) by considering given sample images as seen in the group annotation view of Fig.~\ref{sup-fig:annotation_tool}.
Next, we manually check the validity of a single morph one-by-one by examining its corresponding morph-applied image as shown in individual inspection view of Fig.~\ref{sup-fig:annotation_tool}.
If the morph-applied image has an unmatched semantic, we exclude that source morph marking it as X.
We present the annotation results in Table~\ref{sup-table:mapping}.

\noindent\textbf{Pose Sampling Process.}~\label{subsec:pose_sampling}
Algorithm~\ref{alg:pose_sampling} depicts a detailed process for sampling a pose vector $\mathbf{p} \in \mathbb{R}^{20}$.
Note that the annotated target morphs can be different depending on the 3D animation model.
Given the annotated target morphs $\{e_{n}\}_{n=1}^{N}$, we first select a semantic of each part: eye $\text{s}_{eye}$, eyebrow $\text{s}_{eyebrow}$, and mouth $\text{s}_{mouth}$.
For example, if there exist Mouth (A), Mouth (E) and Mouth (O) as mouth semantics, we randomly sample one of them as $\text{s}_{mouth}$.
Similar to this, the pre-defined target morphs are randomly sampled for $\text{s}_{eye}$ and $\text{s}_{eyebrow}$, respectively. 
The difference is that we check whether a 3D animation model contains independent morphs that can control left and right part separately or a single morph to adjust both parts.
If there exist the independent morphs, they are used with priorities.
Then, the values sampled from a uniform distribution are assigned to the selected semantics as well as head angles (\textit{i.e.}, roll, pitch, and yaw).
This results in a pose vector $\mathbf{p}$ that works for manipulating a pose of an animation character.

\noindent\textbf{Sampling Examples.}
Fig.~\ref{sup-fig:sampling} shows the example pairs of generated images and pose vectors from a 3D animation model.
The output data consists of \textit{frontalized-expression} and \textit{rotated-expression} images and their corresponding pose vectors that contain 17 different morphs and 3D head angles. 
In addition, we provide four different shader styles: (S.1), (S.2), (S.3) and (S.4) to boost the diversity of images and consider various drawing styles of animation creators.

% Figure - Data Sampling

% Morph Mapping Table
\begin{table}[h!]
    \scriptsize
    \centering
    \begin{tabular}{l | l}
        \toprule
        \textit{Source morphs} & \textit{Target morphs}  \\
        \toprule
        あ, ああ, あ2 & Mouth(A)   \\ 
        \hline
        え, ええ, え2, えー & Mouth(E)  \\
        \hline
        い, いい, い2, いー & Mouth(I)  \\ 
        \hline
        お, おお& Mouth(O)  \\ 
        \hline
        う, うう& Mouth(U)   \\ 
        \hline
        ばたき, 笑い, なごみ& Closed Eyes  \\ 
        \hline
        ウィンク, ウィンク.001, ウィンク2, なごみ左 & Left Closed Eye  \\ 
        \hline
        ウインク右, なごみ右, ウインク２右, ｳｨﾝｸ２右 & Right Closed Eye \\ 
        \hline
        半目, じと目, ジト目& Unimpressed Eye  \\ 
        \hline
        じと目左& Left Unimpressed Eye  \\ 
        \hline
        じと目右& Right Unimpressed Eye  \\ 
        \hline
        びっくり２, びっくり, 驚き & Surprised Eyes \\
        \hline
        びっくり左, びっくり２左 & Left Surprised Eye \\
        \hline
        びっくり２右, びっくり右 & Right Surprised Eye \\
        \hline
        怒り眉, 怒り2, 怒り & Angry Eyebrows \\
        \hline
        怒り左, '怒り眉左, 怒りL & Left Angry Eyebrow \\
        \hline
        怒り眉右, 怒り右, 怒りR & Right Angry Eyebrow \\
        \hline
        上 & Raised Eyebrows \\
        \hline
        上左, 上L & Left Raised Eyebrow \\
        \hline
        上右, 上R & Right Raised Eyebrow \\
        \hline
        下, 困る & Lowered Eyebrows \\
        \hline
        困るL, 下L, 困る左, 下左 & Left Lowered Eyebrow \\
        \hline
        困る右, '下R, 下右, 困るR & Right Lowered Eyebrow \\
        \bottomrule
    \end{tabular}
    \caption{Examples of mapping relationships between the source morphs and the target morphs.}
    \vspace{-0.5cm}
    \label{sup-table:mapping}
\end{table}

% Figure - Annotation Tools

% Pseudo Codes for Pose Sampling
\RestyleAlgo{ruled} % Change the Algorithm Style
\SetKwComment{Comment}{/* }{ */} % Comment

\begin{algorithm}[t!]
\caption{Pseudo Codes for Pose Sampling}\label{alg:pose_sampling}
\small
\KwData{Annotated target morphs $\{e_{n}\}_{n=1}^{N}$} \Comment{N indicates the number of source morphs of a 3D animation model.}
\KwResult{A sampled pose $\mathbf{p} \in \mathbb{R}^{20}$}

\Comment{Select eye, eyebrow, mouth semantics and sample the values from a uniform distribution.}
$\text{s}_{eye},\text{s}_{eyebrow},\text{s}_{mouth}$ $\leftarrow$ sample($\{e_{n}\}$)\\
\eIf{$\exists \: left\text{-s}_{eye},right\text{-s}_{eye} \in \{e_{n}\}$}{
  $u_1, u_2 \sim \mathcal{U}(0,1)$\;
  $left\text{-s}_{eye}(v) \leftarrow u_1$\;
  $right\text{-s}_{eye}(v) \leftarrow u_2$\;
}{
  $u \sim \mathcal{U}(0,1)$\;
  $left\text{-s}_{eye}(v) \leftarrow u$\;
  $right\text{-s}_{eye}(v) \leftarrow u$\;
}
\eIf{$\exists \: left\text{-s}_{eyebrow},right\text{-s}_{eyebrow} \in \{e_{n}\}$}{
  $u_1, u_2 \sim \mathcal{U}(0,1)$\;
  $left\text{-s}_{eyebrow}(v) \leftarrow u_1$\;
  $right\text{-s}_{eyebrow}(v) \leftarrow u_2$\;
}{
  $u \sim \mathcal{U}(0,1)$\;
  $left\text{-s}_{eyebrow}(v) \leftarrow u$\;
  $right\text{-s}_{eyebrow}(v) \leftarrow u$\;
}
$u \sim \mathcal{U}(0,1)$\;
$\text{s}_{mouth}(v) \leftarrow u$\;

\Comment{Sample roll, pitch and yaw from a uniform distribution.}
$\text{roll}(v),\text{pitch}(v),\text{yaw}(v) \sim \mathcal{U}(-20^{\circ},20^{\circ})$\;

\Comment{Fill $\mathbf{p}$ with sampled values. $\mathbf{p}[\cdot]$ denotes an index of each semantic.}
initialize $\mathbf{p} = \{p_{m}\}_{m=1}^{20} = \{0,0,...,0\}$\;

$\mathbf{p}[left\text{-s}_{eye}] = left\text{-s}_{eye}(v)$\;
$\mathbf{p}[right\text{-s}_{eye}] = right\text{-s}_{eye}(v)$\;
$\mathbf{p}[left\text{-s}_{eyebrow}] = left\text{-s}_{eyebrow}(v)$\;
$\mathbf{p}[right\text{-s}_{eyebrow}] = right\text{-s}_{eyebrow}(v)$\;
$\mathbf{p}[\text{s}_{mouth}] = \text{s}_{mouth}(v)$\;
$\mathbf{p}[\text{roll}] = \text{roll}(v)$\;
$\mathbf{p}[\text{pitch}] = \text{pitch}(v)$\;
$\mathbf{p}[\text{yaw}] = \text{yaw}(v)$\;
\end{algorithm}
\clearpage
\section{Additional AnimeCeleb Samples and Experimental Results}~\label{animeceleb_additional}

This section presents additional results as follows:
\begin{itemize}
    \item Other examples sampled from the AnimeCeleb.
    \item Qualitative head reenactment results on other animation head images obtained from Waifu Labs~\footnote{https://waifulabs.com/} and Danbooru 2019~\cite{danbooru2019Figures}.
    \item Other applicable tasks using the AnimeCeleb: animation colorization and image harmonization.
\end{itemize}

\begin{figure}[h!]
    \centering
    \includegraphics[width=1.0\linewidth]{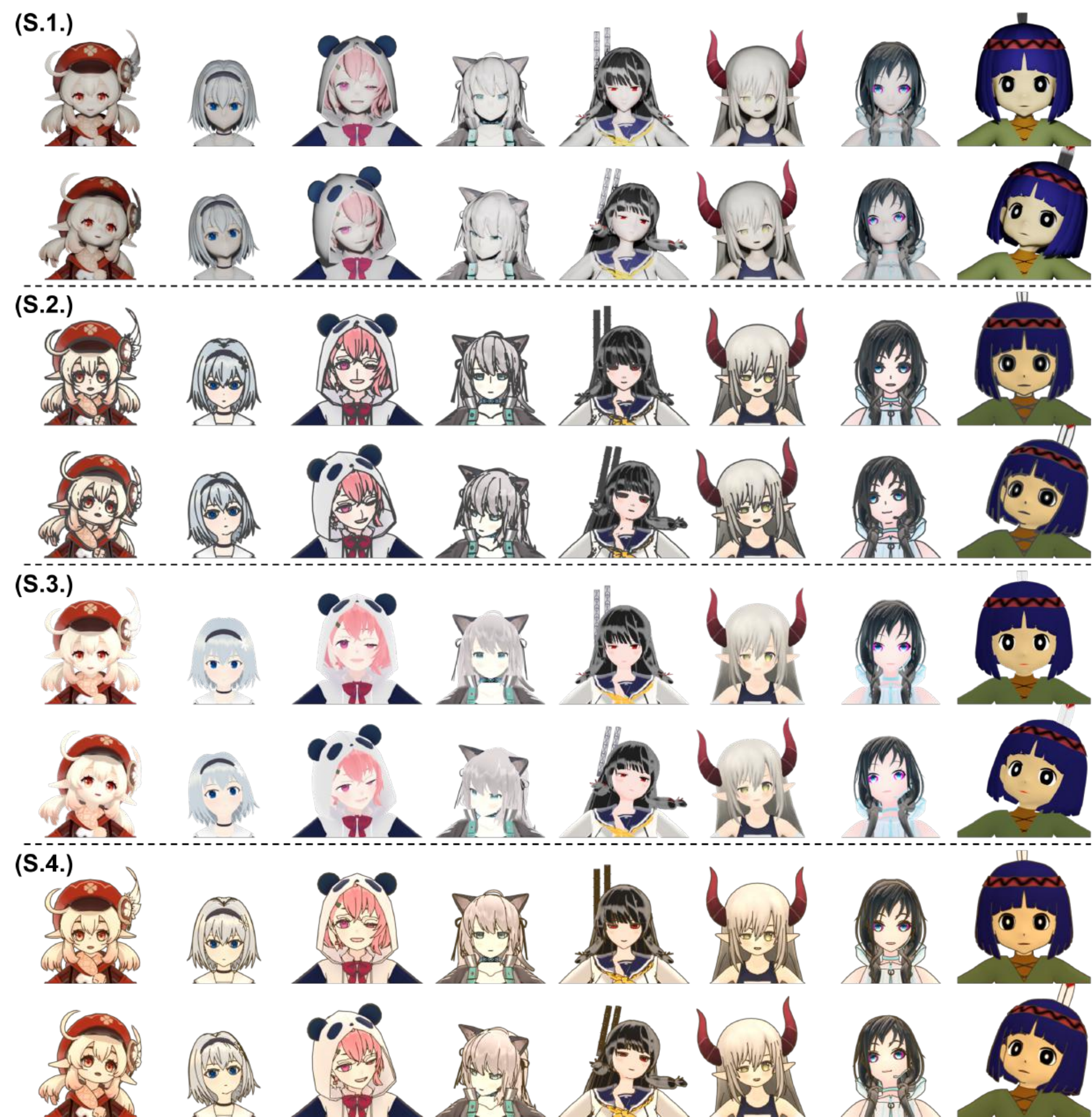}
    \caption{Examples of the created images from the AnimeCeleb.}
    \label{sup-fig:data}
\end{figure}

\noindent\textbf{Additional Examples from AnimeCeleb.}
Fig.~\ref{sup-fig:data} shows the sampled images of eight different characters.
As aforementioned, we present two image groups: \textit{frontalized-expression} (the first row) and \textit{rotated-expression} (the second row), and a difference between two groups lies in whether a head rotation is applied to the animation heads or not.
As seen in Fig.~\ref{sup-fig:data}, the images rendered with different shaders are generated with the exact same pose vector ((S.2-4) in Fig.~\ref{sup-fig:data}) for the purpose of providing multiple styles of images.

\begin{figure}[h!]
    \centering
    \includegraphics[width=0.9\linewidth]{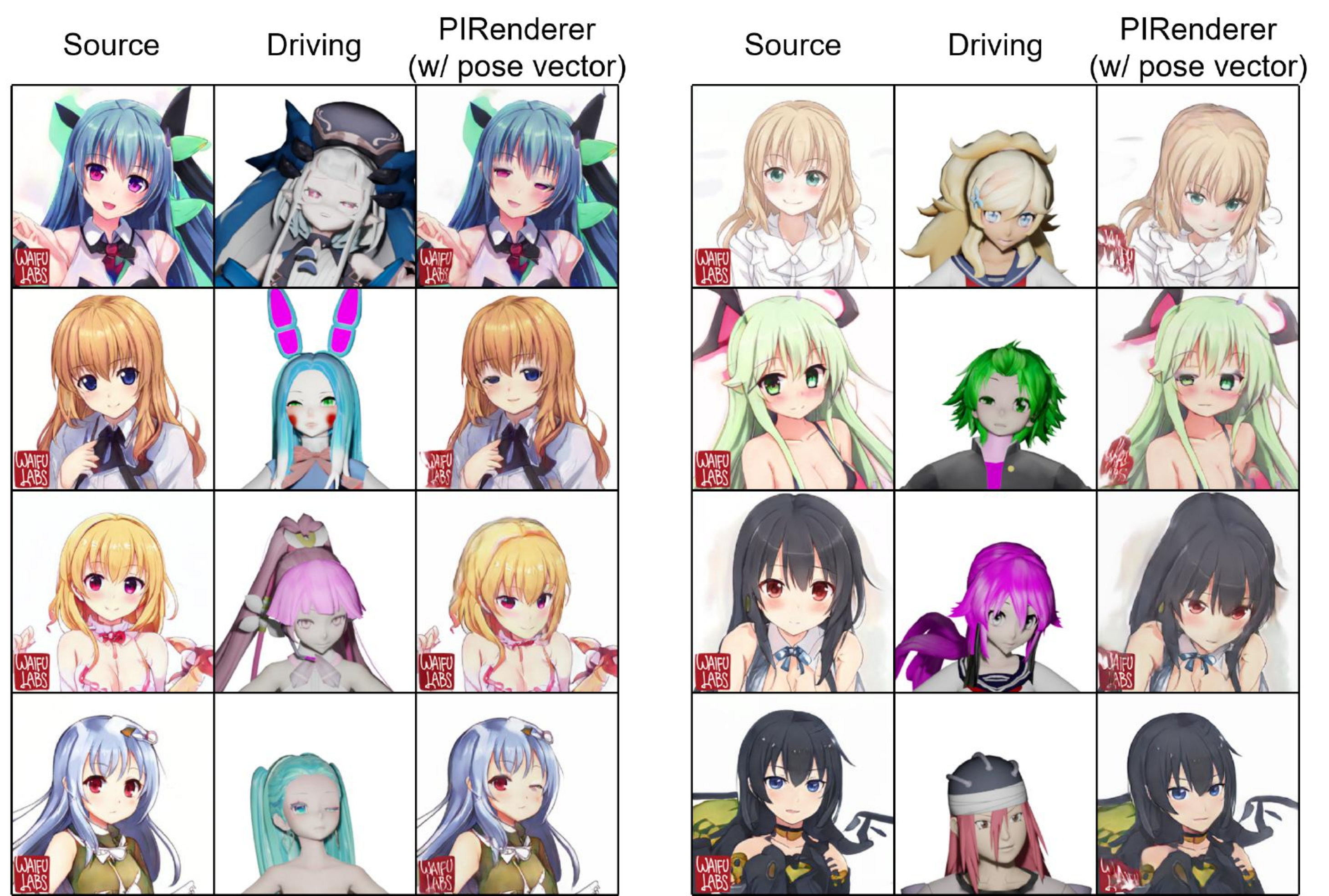}
    \vspace{-0.2cm}
    \caption{Additional animation head reenactment results on the images from Waifu Labs.}
    \label{sup-fig:out-of-domain-waifu}
    
    \centering
    \includegraphics[width=0.9\linewidth]{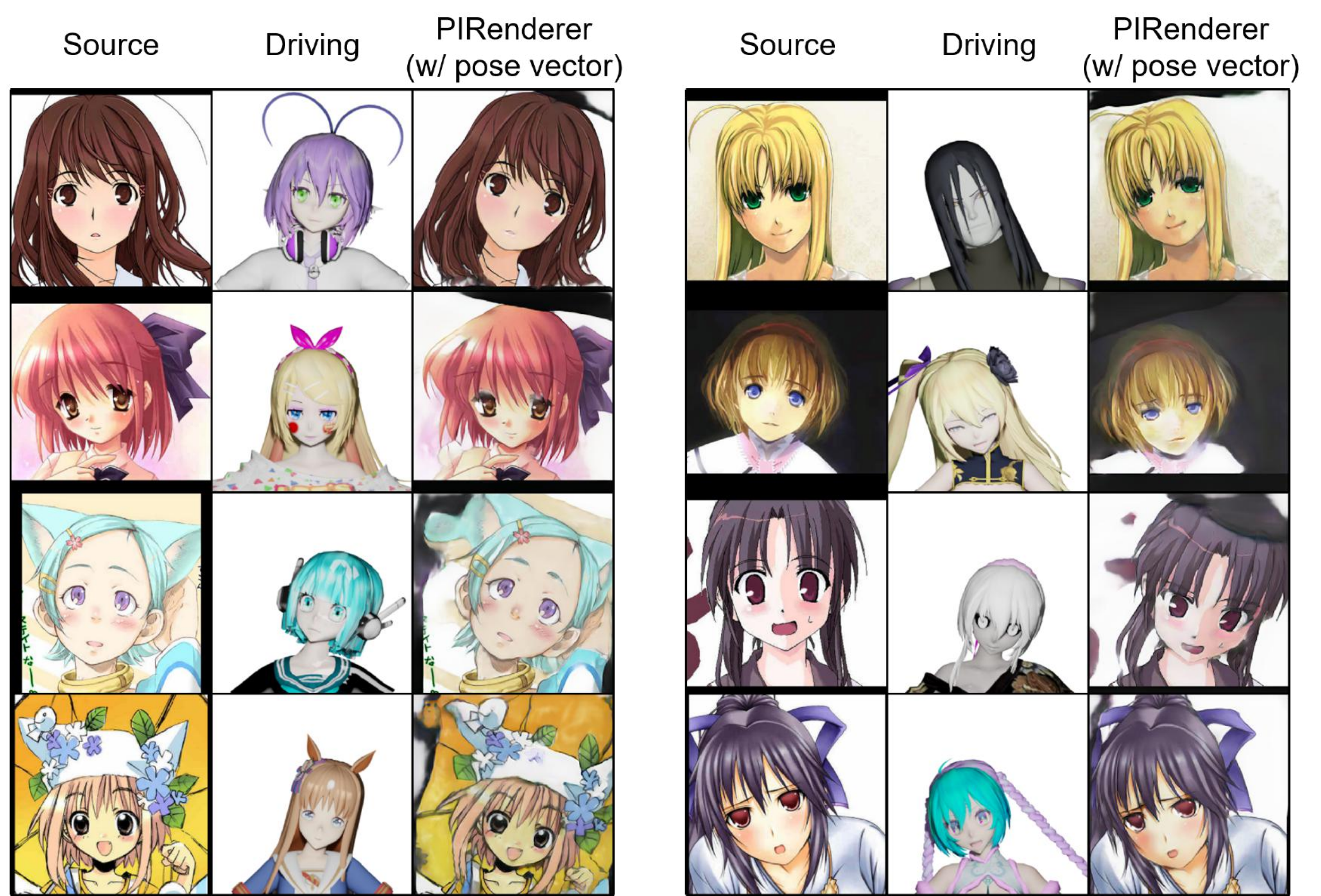}
    \vspace{-0.2cm}
    \caption{Additional animation head reenactment results on the Danbooru 2019.}
    \label{sup-fig:out-of-domain-danbooru}
\end{figure}

\noindent\textbf{Other Animation Images Head Reenactment.}
We present qualitative results using the PIRenderer~\cite{ren2021pirenderer} trained with the AnimeCeleb on two other animation sample images obtained from the Waifu Labs and the Danbooru 2019. 
We choose to use the PIRenderer(w/ pose vector) because it has strong generalization capacity compared to other models as seen in main manuscript.
As shown in Fig.~\ref{sup-fig:out-of-domain-waifu}, the trained model successfully generates the head reenactment results given a source and a driving image.
The PIRenderer(w/ pose vector) produces favorable outputs, imitating the head poses of driving images.
Furthermore, Fig.~\ref{sup-fig:out-of-domain-danbooru} shows the outcomes on the Danbooru 2019.
Due to the distribution gap between the AnimeCeleb and the Danbooru 2019, we confirm slight performance degradation for the samples from the Danbooru 2019.

\noindent\textbf{Additional Applications of AnimeCeleb.}
To reveal the benefits of the AnimeCeleb, we implement additional two tasks: an \textit{animation colorization}, and an \textit{image harmonization}.
The third shader (\textit{i.e.}, S.3) styled images are used to train the colorization and the harmonization models.
We clarify an importance of each task in the animation domain and show experimental results in the following paragraphs. 

\begin{figure}[t!]
    \centering
    \includegraphics[width=1.0\linewidth]{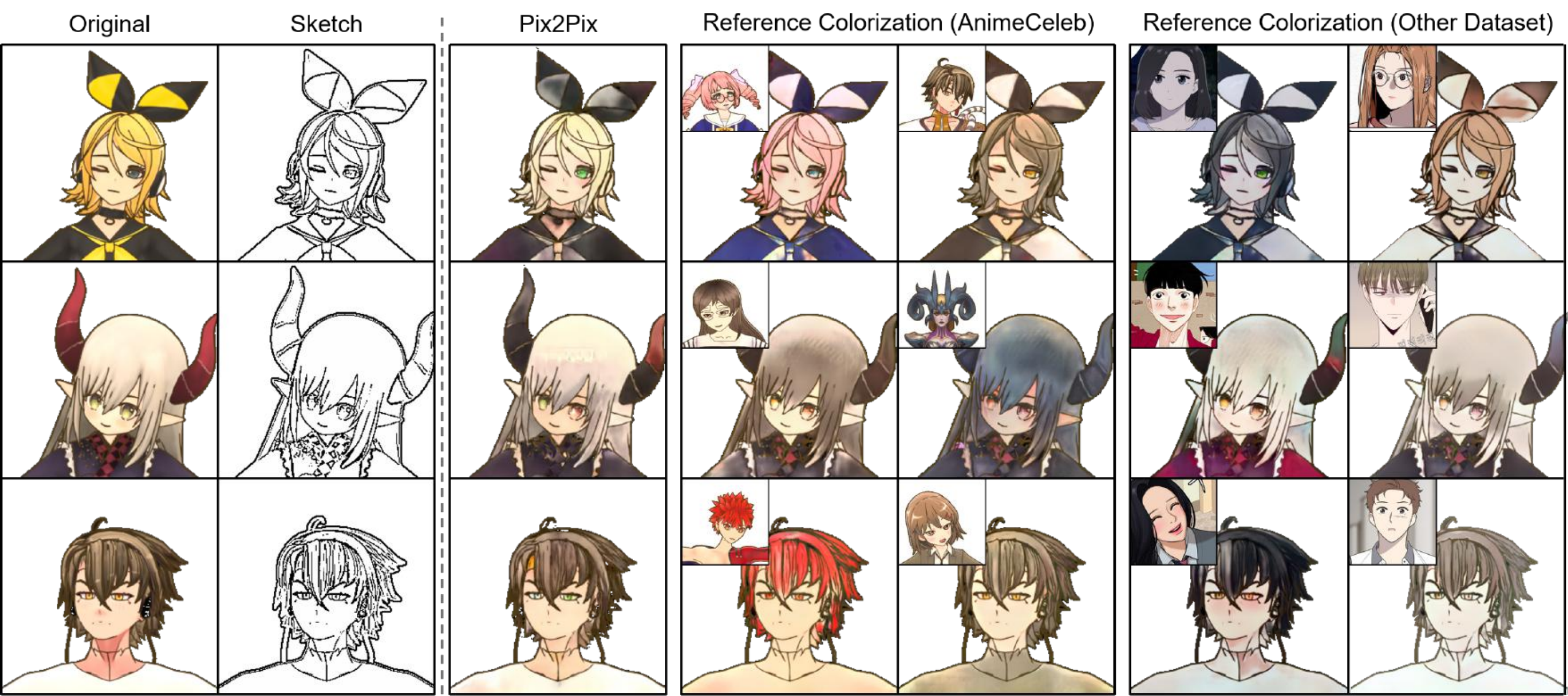}
    \caption{Colorization results in an automatic and reference-based manner on the AnimeCeleb and other collected images.
    A Pix2Pix~\cite{isola2017image} trained with the AnimeCeleb successfully outputs a plausible colorized image.
    Also, a reference-based model~\cite{referencecolor} successfully fills a given sketch image with the color maps extracted from reference images.}
    \label{sup-fig:colorization}
\end{figure}

First, the animation colorization is a practical task for animation creators to reduce their effort during the labor-intensive painting process.
Given a trained colorization model, creators are able to obtain colorized images given sketch images.
We conduct character colorization tasks using both unconditional and conditional colorization baselines~\cite{isola2017image,lee2020reference}.
As can be seen in Fig.~\ref{sup-fig:colorization}, the colorization models trained with the AnimeCeleb show a promising performance at painting the animation character sketch images, producing plausible colorization outputs in an automatic manner or following a given animation reference image.
To demonstrate the broad generalization capacity of the reference-based colorization model~\cite{lee2020reference} trained with the AnimeCeleb, we also use the reference images crawled from online cartoons.
We find that not limited to the AnimeCeleb reference images, the model achieves high-quality colorization outputs based on other animation head images.

\begin{figure}[t!]
    \centering
    \includegraphics[width=0.6\linewidth]{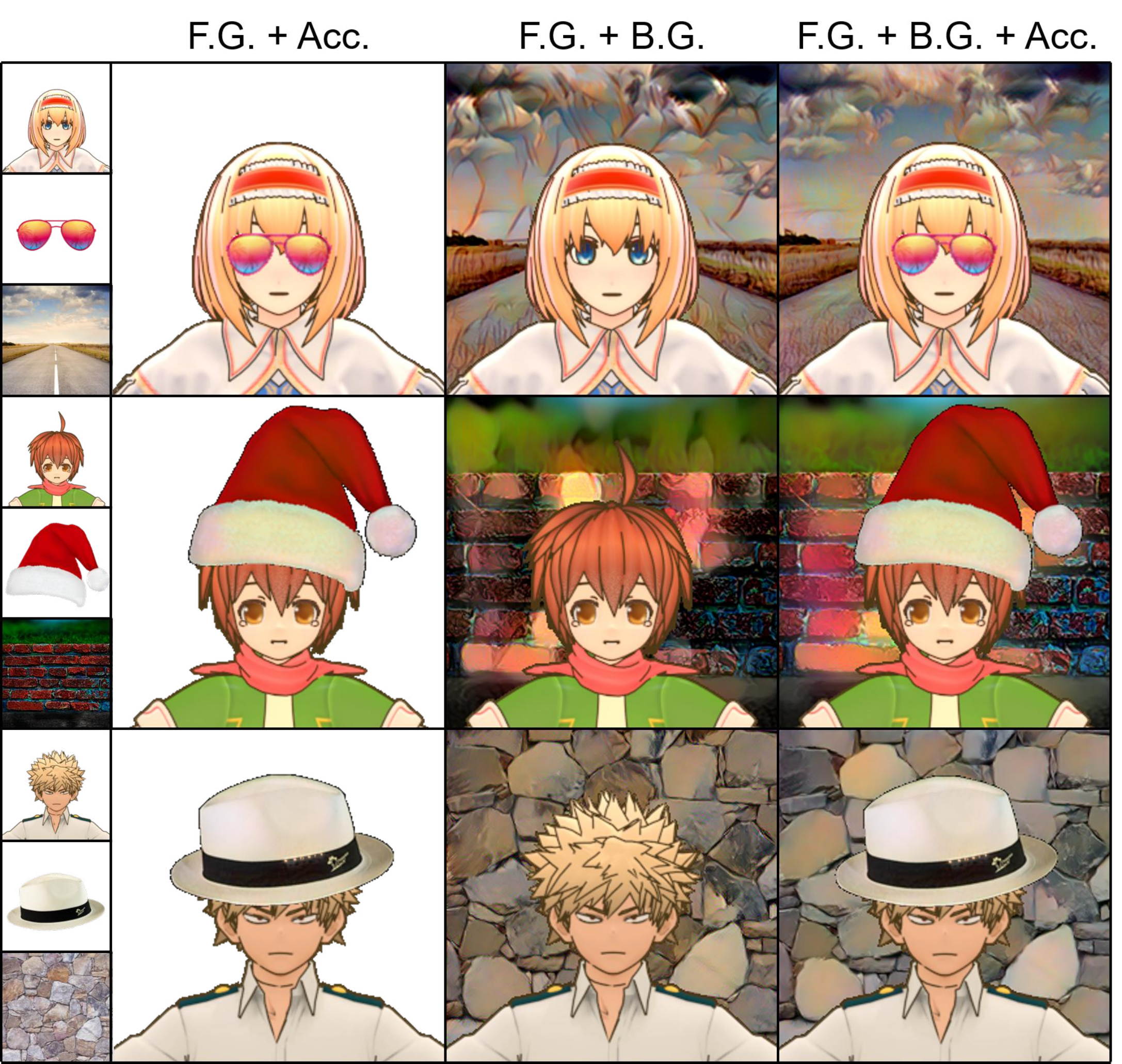}
    \caption{Image harmonization results. F.G., B.G. and Acc. denotes a foreground object, a background, and an accessory, respectively. The components for image harmonization (the \textit{1st column}) are well-blended, where the backgrounds and the accessories are refined with similar styles with an animation character.}
    \label{sup-fig:harmonization}
\end{figure}

Second, the image harmonization aims to generate natural composite images given two images from different domains, achieving a visually pleasing match for both content and style.
We implement a representative optimization-based approach~\cite{zhang2020deep} to explore the applicability of the AnimeCeleb and generate more realistic animation images.
Since the AnimeCeleb images only contain a foreground object (\textit{i.e.}, an animation head), a composition with suitable background is a natural extension of the AnimeCeleb.
Not limited to the background composition, decorative objects (\textit{e.g.}, sunglasses, caps and masks) are available assets to be exploited for the composition.
We can employ an optimization-based composition model~\cite{zhang2020deep} that requires a foreground segmentation mask because the AnimeCeleb includes the segmentation mask.
As shown in Fig.~\ref{sup-fig:harmonization}, both background and decorative object composition with the AnimeCeleb produce plausible results, demonstrating a potential extension of the AnimeCeleb in that it can provide the images with diverse backgrounds and multiple objects.

% Figure - Data Samples

% Figure - Animation Out-Domain Face Reenactment Results (Waifu,Danbooru)

% Figure - Animation Colorization Results

% Figure - Animation Harmonization Results

\clearpage
\section{Implementation Details of the \model and Baselines}~\label{animo_details}
In this section, we describe the architectures of the motion network, the warping network, and the editing network in detail, and objective functions for training.
Then, we elaborate the baselines~\cite{firstorder,ren2021pirenderer,latentpose} and implementation details of them, respectively.

% Figure - Model Details: Mapping Network
\begin{figure}[h!]
    \centering
    \includegraphics[width=0.45\linewidth]{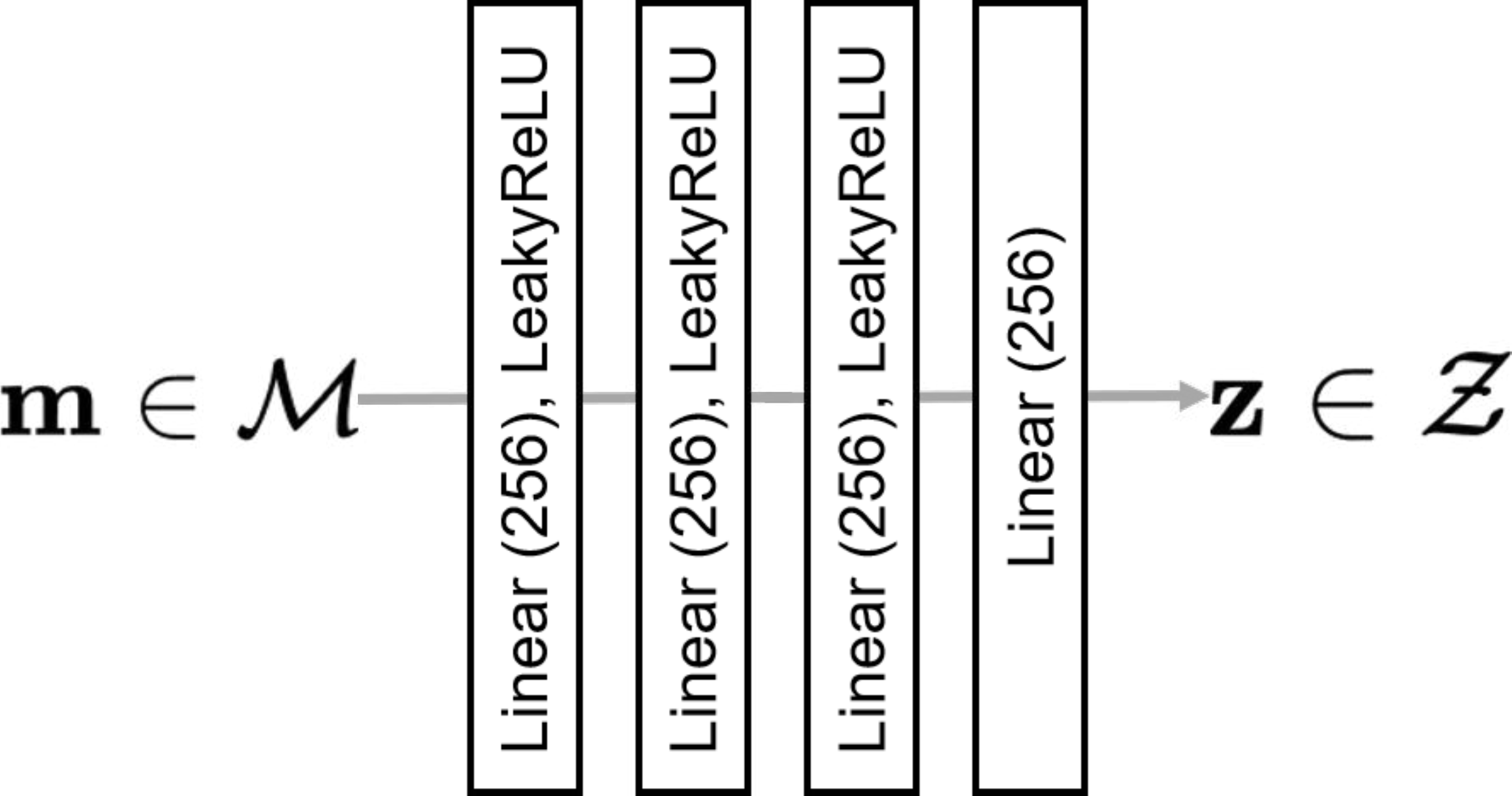}
    \caption{The architecture of the motion network.}
    \label{supp-fig:motion_net}
\end{figure}

\noindent\textbf{Motion Network.}
As shown in Fig.~\ref{supp-fig:motion_net}, the motion network has a multi-layer perceptron structure, which consists of four fully-connected layers that are responsible for resulting in a latent motion code $\mathbf{z} \in \mathbb{R}^{256}$ given the 3DMM parameters $\mathbf{m} \in \mathbb{R}^{70}$.
The latent motion code $\mathbf{z}$ are transformed to estimate the affine parameters for adaptive instance normalization (AdaIN)~\cite{huang2017arbitrary} operations in the warping network and the editing network.

% Figure - Model Details: Flow Predictor
\begin{figure}[h!]
    \centering
    \includegraphics[width=0.75\linewidth]{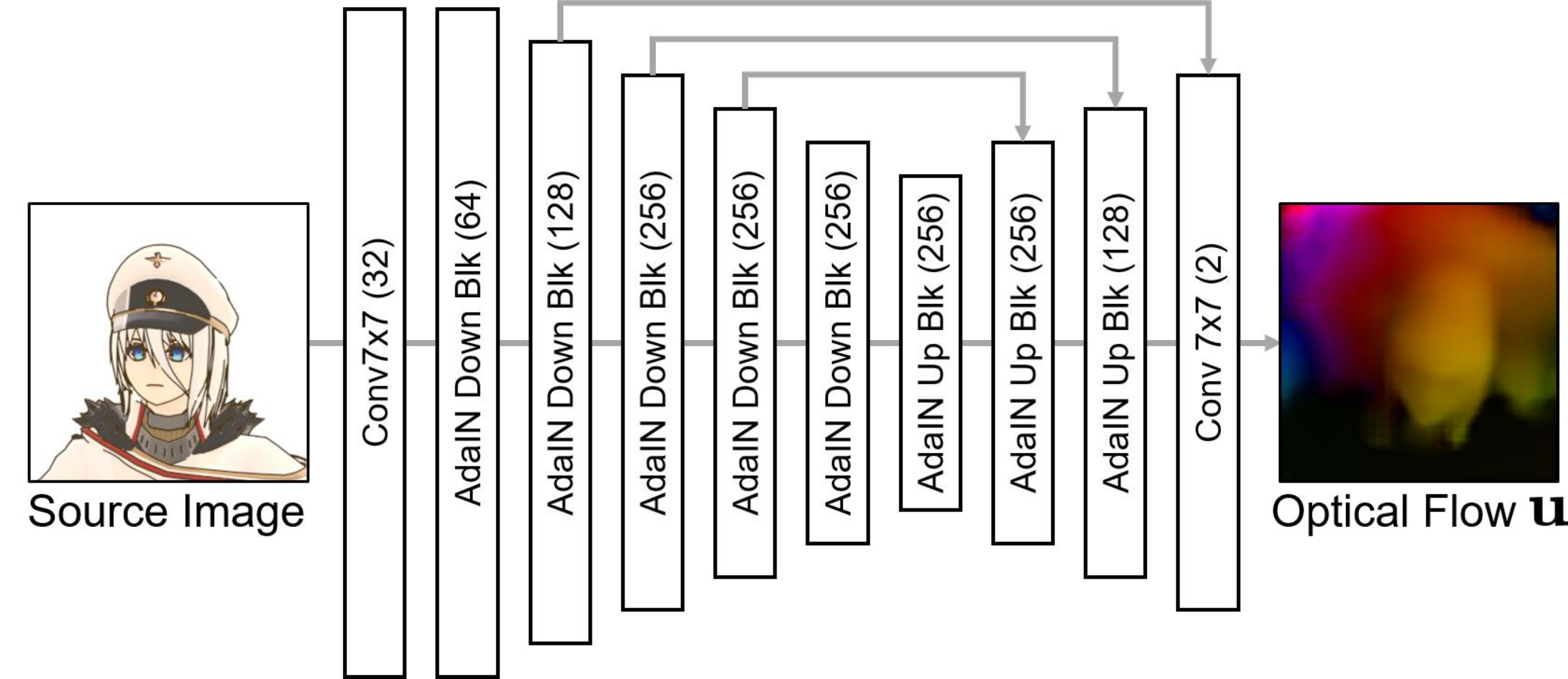}
    \caption{The architecture of the warping network.}
    \label{supp-fig:flow_predictor}
\end{figure}

\noindent\textbf{Warping Network.}
As shown in Fig.~\ref{supp-fig:flow_predictor}, the warping network has a encoder-decoder architecture. 
In addition, we employ the skip-connection as U-Net~\cite{ronneberger2015u} to preserve the spatial information as well as AdaIN operation to inject the motion information.
The optical flow $\mathbf{u} \in \mathbb{R}^{64 \times 64 \times 2}$ is upsampled or downsampled to fit the sizes of feature maps in the editing network.

% Figure - Model Details: Image Generator
\begin{figure}[h!]
    \centering
    \includegraphics[width=0.75\linewidth]{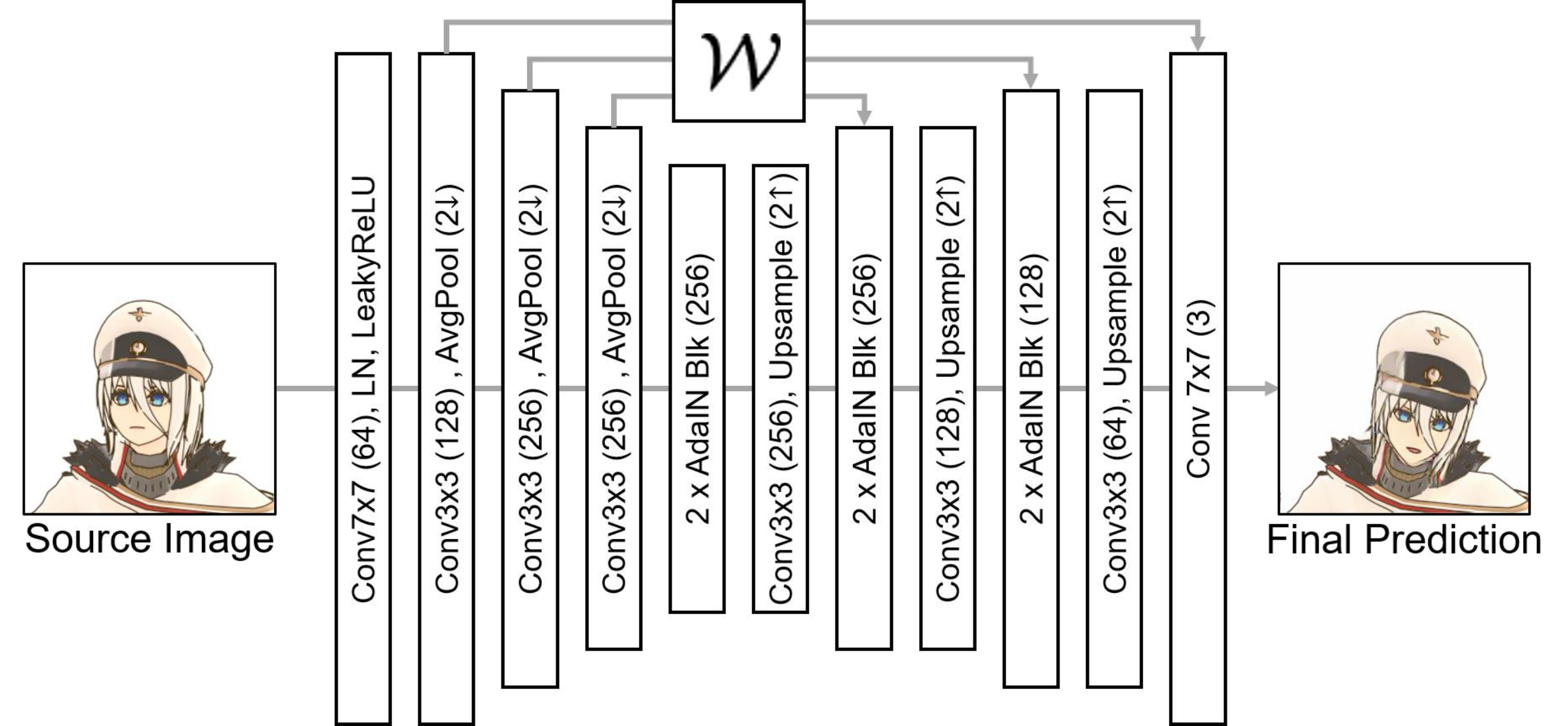}
    \caption{The architecture of the editing network.}
    \label{supp-fig:image_generator}
\end{figure}

\noindent\textbf{Editing Network.}
Fig.~\ref{supp-fig:image_generator} shows the architecture of the editing network.
The editing network employs the structure of a hourglass network~\cite{newell2016stacked}, in which intermediate encoder feature maps are passed to the decoder layers by an element-wise addition operation.
When propagating the multi-scale feature maps of the encoder to the decoder, the optical flow $\mathbf{u}$ is applied to the multi-scale feature maps.
In addition, as similar to the warping network, we utilize the AdaIN operation to inject the motion information.

\noindent\textbf{Objective Functions.}
In order to train the \model, we follow the PIRenderer~\cite{ren2021pirenderer} objective functions as follows.

First, a reconstruction loss encourages the warping network to estimate an accurate optical flow.
For the sake of this, we apply the estimated optical flow to a source image $s$, and encourage the warped output to reconstruct a driving image $d$.
Instead of pixel-level loss, we employ the perceptual loss~\cite{johnson2016perceptual} to minimize the $\ell_{1}$ distances in latent feature space between the warped source image $\mathbf{u}(s)$ and driving image $d$.
Formally, this can be written as:
\begin{equation}
    \mathcal{L}^{w}_{perc}(s, d) = \sum_{j}\norm\Big{\phi_{j}(\mathcal{W}(s, \mathbf{u})) - \phi_{j}(d)} _{1},
\end{equation}
where $\phi_{j}$ represents the activation map of $j$-th layer of the pre-trained VGG-19 network~\cite{simonyan2014very} and $\mathcal{W}$ denotes a warping operation.
This leads to reliable optical flow prediction of the warping network.

Second, our editing network is trained with two losses: a reconstruction loss $\mathcal{L}^{g}_{perc}$ and a style loss $\mathcal{L}^{g}_{sty}$.
The reconstruction loss is designed to reduce the errors between the final prediction $\hat{d}$ and the ground-truth driving image $d$.
This can be formulated as:
\begin{equation}
    \mathcal{L}^{g}_{perc}(d, \hat{d}) = \sum_{j}\norm\Big{\phi_{j}(\hat{d}) - \phi_{j}(d)} _{1}.
\end{equation}
Next, the style loss is introduced to match the statistics between the ground truth driving image $d$ and the final prediction as follows:
\begin{equation}
    \mathcal{L}^{g}_{sty}(d, \hat{d}) = \sum_{j}\norm\Big{C^{\phi}_{j}(\hat{d}) - C^{\phi}{j}(d)} _{1},
\end{equation}
where $C^{\phi}_{j}$ denotes the gram matrix calculated from the activation maps $\phi_{j}$.

In summary, our full objective function is given as:
\begin{align*}
    \mathcal{L}_{total} &= 
        \lambda^{w}_{perc}(\mathcal{L}^{w}_{perc}(s^{(a)}, d^{(a)}) + \mathcal{L}^{w}_{perc}(s^{(r)}, d^{(r)})) \\
         &+ \lambda^{g}_{perc}(\mathcal{L}^{g}_{perc}(d^{(a)}, \hat{d}^{(a)}) + \mathcal{L}^{g}_{perc}(d^{(r)}, \hat{d}^{(r)})) \\
         &+ \lambda^{g}_{sty}(\mathcal{L}^{g}_{sty}(d^{(a)}, \hat{d}^{(a)}) + \mathcal{L}^{g}_{sty}(d^{(r)}, \hat{d}^{(r)})),
\end{align*}
where $\lambda^{w}_{perc}, \lambda^{g}_{perc}$ and $\lambda^{g}_{sty}$ are hyperparameters that control the relative importance of three different losses.
We set $\lambda^{w}_{perc}, \lambda^{g}_{perc}$ and $\lambda^{g}_{sty}$ as 2.5, 4 and 250, respectively.
Note that our framework is jointly trained on both the AnimeCeleb and VoxCeleb. 

We train the \model in two stages, where the motion network and the warping network are trained for 100 epochs, and we train the entire network for the additional 100 epochs.
We employ the Adam~\cite{kingma2014adam} optimizer, one of the widely-used optimization methods, with the learning rate of 0.0001.
The learning rate is set initially as 0.0001, then decreased to 0.00002 after 150 epochs.
The batch size is set to 12 for all experiments.

\noindent\textbf{Head Reenactment Baselines.}
We compare the \model with state-of-the-art models~\cite{latentpose,firstorder,ren2021pirenderer}.
Since we leverage two datasets during training, comparable baselines are trained on either the VoxCeleb following their original implementations or both the VoxCeleb and AnimeCeleb.

In the following, we describe each baseline and experimental settings:
\begin{itemize}
    \item \textbf{First-Order Motion Model (FOMM)}~\cite{firstorder} is an unsupervised landmark-based approach, which internally detects the spatial positions to transform the source image. 
    We implement two versions of this model: a VoxCeleb-trained and a jointly-trained model using both the AnimeCeleb and the VoxCeleb.
    \item \textbf{PIRenderer}~\cite{ren2021pirenderer} takes the 3DMM parameters to represent a driving motion and employs the AdaIN operation to inject the motion information.
    Similar to FOMM, we first implement a VoxCeleb-trained model. 
    Also, we apply our pose mapping $\mathcal{T}$ to use a shared pose representations (\textit{i.e.}, 3DMM parameters) for the purpose of achieving joint training.
    \item \textbf{Latent Pose Descriptor (LPD)~\cite{latentpose}} relies on the AdaIN operation to inject a motion information, where the driving image is encoded as latent pose vector in unsupervised manner.
    To handle an unseen identity during inference, a trained model is fine-tuned with the same-identity images to infer.
    For evaluation, we utilize a model trained on the VoxCeleb, and fine-tune it using a group of the same-identity images in the AnimeCeleb.
\end{itemize}

For the implementations of existing baselines, we follow the hyper-parameters given in the original papers and codes.

\clearpage
\section{Additional Head Reenactment Results of \textit{AniMo}}~\label{animo_additional}
\vspace{-0.4cm}

This section contains additional head reenactment results with the \model and the baselines as follows:
\begin{itemize}
    \item Qualitative results on self-identity (VoxCeleb and AnimeCeleb), cross-identity (VoxCeleb and AnimeCeleb), and cross-domain head reenactment (Vox. $\rightarrow$ Anime. and Anime. $\rightarrow$ Vox.) tasks.
    \item Intuitive pose editing of an animation and human head images.
    \item Qualitative results on cross-domain head reenactment using various unseen head images.
    \item A user study to compare the characteristics with iCartoon and head angle distribution comparison between VoxCeleb.
\end{itemize}

\noindent\textbf{Additional Qualitative Head Reenactment Results of \textit{AniMo}.}
% Overview
In the experiments, we utilize two different training source: single dataset (VoxCeleb) and joint datasets (AnimeCeleb and VoxCeleb).
We use the single dataset (VoxCeleb) to compare the original experimental setup of the previous studies~\cite{latentpose,firstorder,ren2021pirenderer}.
For qualitative comparisons, we show the results of three tasks: (1) \textbf{self-identity head reenactment} where the identical being provides both a source and a driving image, (2) \textbf{cross-identity head reenactment} where the identities of a source and driving image are different within the same dataset, and (3) \textbf{cross-domain head reenactment} where two frames of different identities sampled from the AnimeCeleb and the VoxCeleb alternatively for the sake of serving as a source and a driving image; for example, Vox. $\rightarrow$ Anime. denotes a source and driving image are sampled from the AnimeCeleb and the VoxCeleb, respectively.
Note the warping and the editing network for each domain: $W_A, G_A$ and $W_V, G_V$ are responsible for producing an animation and a real human head image, respectively.

\begin{figure}[t!]
    \centering
    \includegraphics[width=1.0\linewidth]{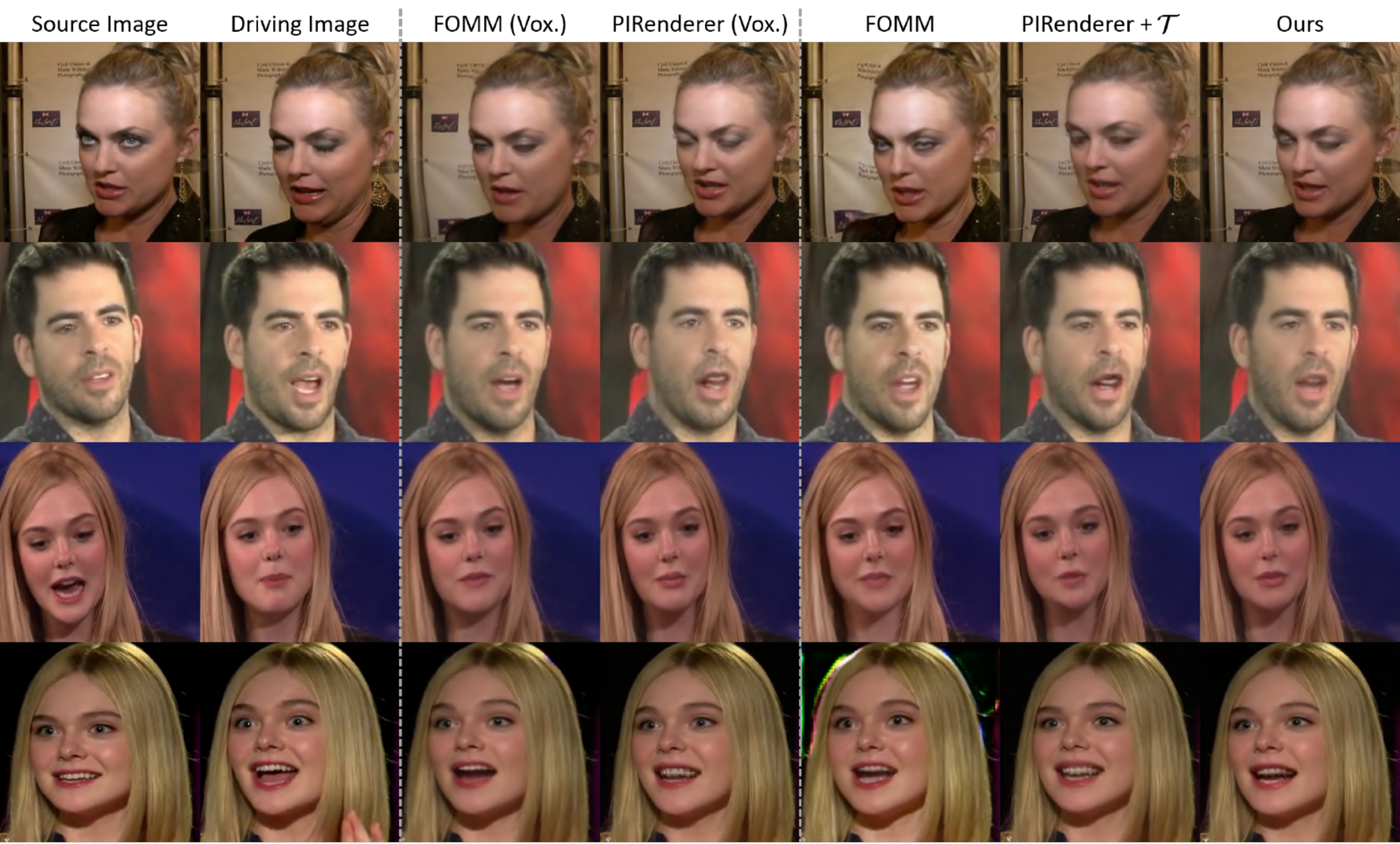}
    \caption{Qualitative comparison between our model and the baselines on self-identity head reenactment given the images of the Voxceleb.}
    \label{supp-fig:baselines-self-identity-vox}
\end{figure}

\begin{figure}[t!]
    \centering
    \includegraphics[width=1.0\linewidth]{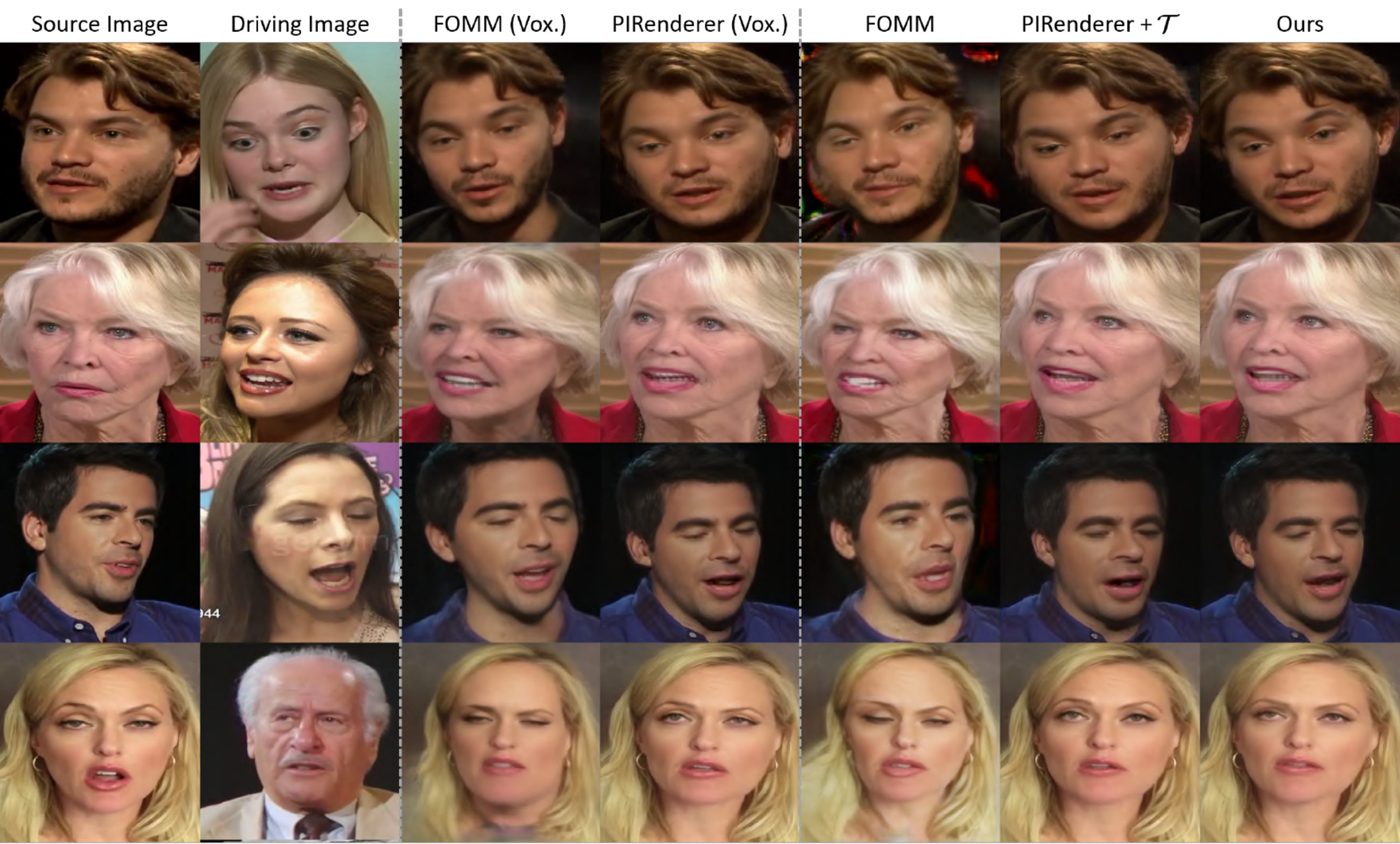}
    \caption{Qualitative comparison between our model and the baselines on cross-identity head reenactment given the images of the Voxceleb.}
    \label{supp-fig:baselines-cross-identity-vox}
\end{figure}

Fig.~\ref{supp-fig:baselines-self-identity-vox} shows qualitative comparisons on self-identity head reenactment using the VoxCeleb.
As seen in Fig.~\ref{supp-fig:baselines-self-identity-vox}, our model produces the outputs that are perceptually realistic, as good as the baselines.
Although the baselines show similar results on the task, there is a performance gap between the models when it comes to handling cross-identity inputs.
As shown in Fig.~\ref{supp-fig:baselines-cross-identity-vox}, the FOMM~\cite{firstorder} often fails to produce photo-realistic results because a head structure of a driving image is involved to generate results (the \textit{3rd} and the \textit{5th} columns).
Compared to these results, the models which rely on the 3DMM parameters successfully handle cross-identity inputs (the \textit{4th}, the \textit{6th} and the \textit{last} columns in Fig.~\ref{supp-fig:baselines-cross-identity-vox}).

\begin{figure}[t!]
    \centering
    \includegraphics[width=1.0\linewidth]{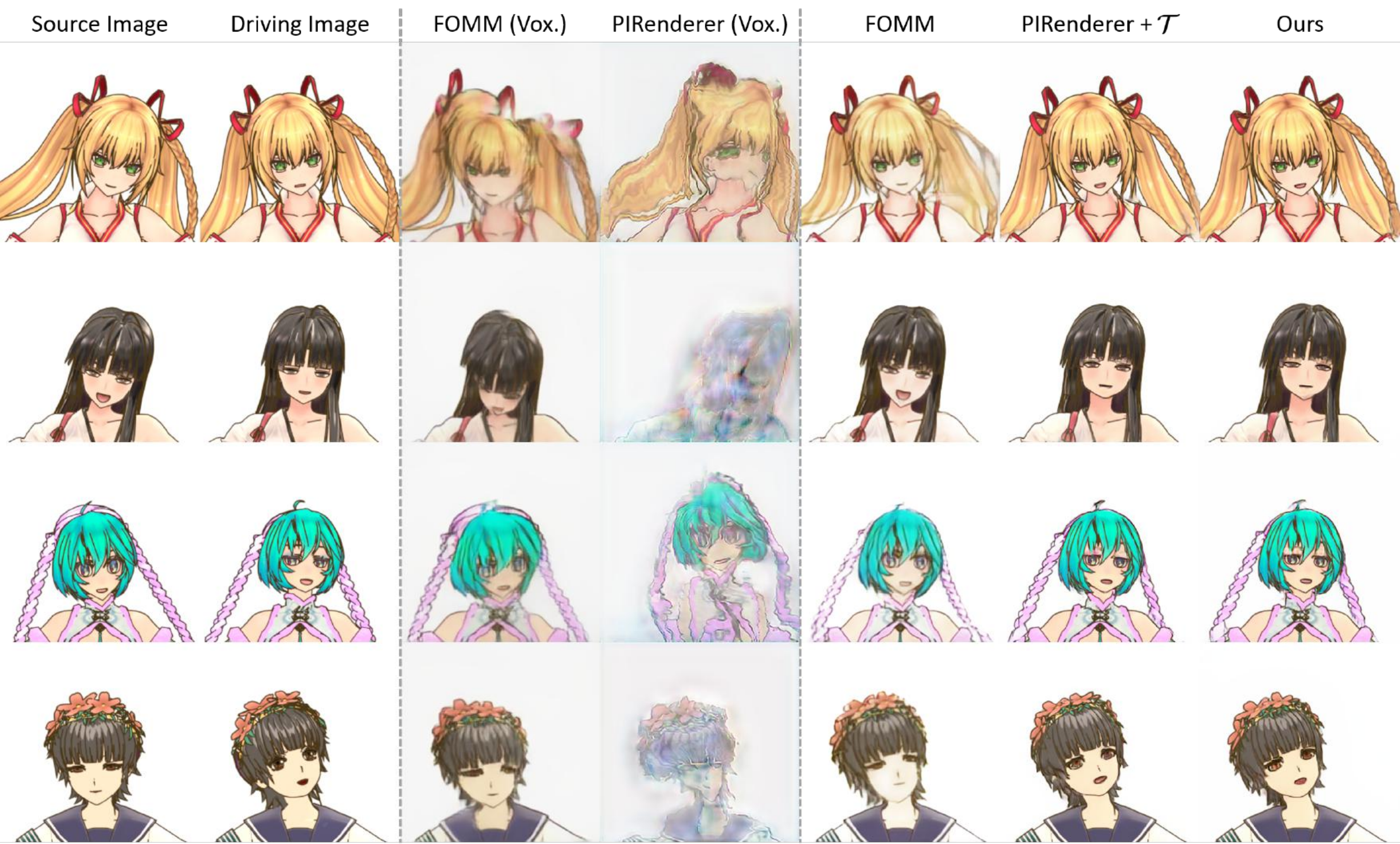}
    \caption{Qualitative comparison between our model and the baselines on self-identity head reenactment given the images of the AnimeCeleb.}
    \label{supp-fig:baselines-self-identity-anime}
\end{figure}

\begin{figure}[t!]
    \centering
    \includegraphics[width=1.0\linewidth]{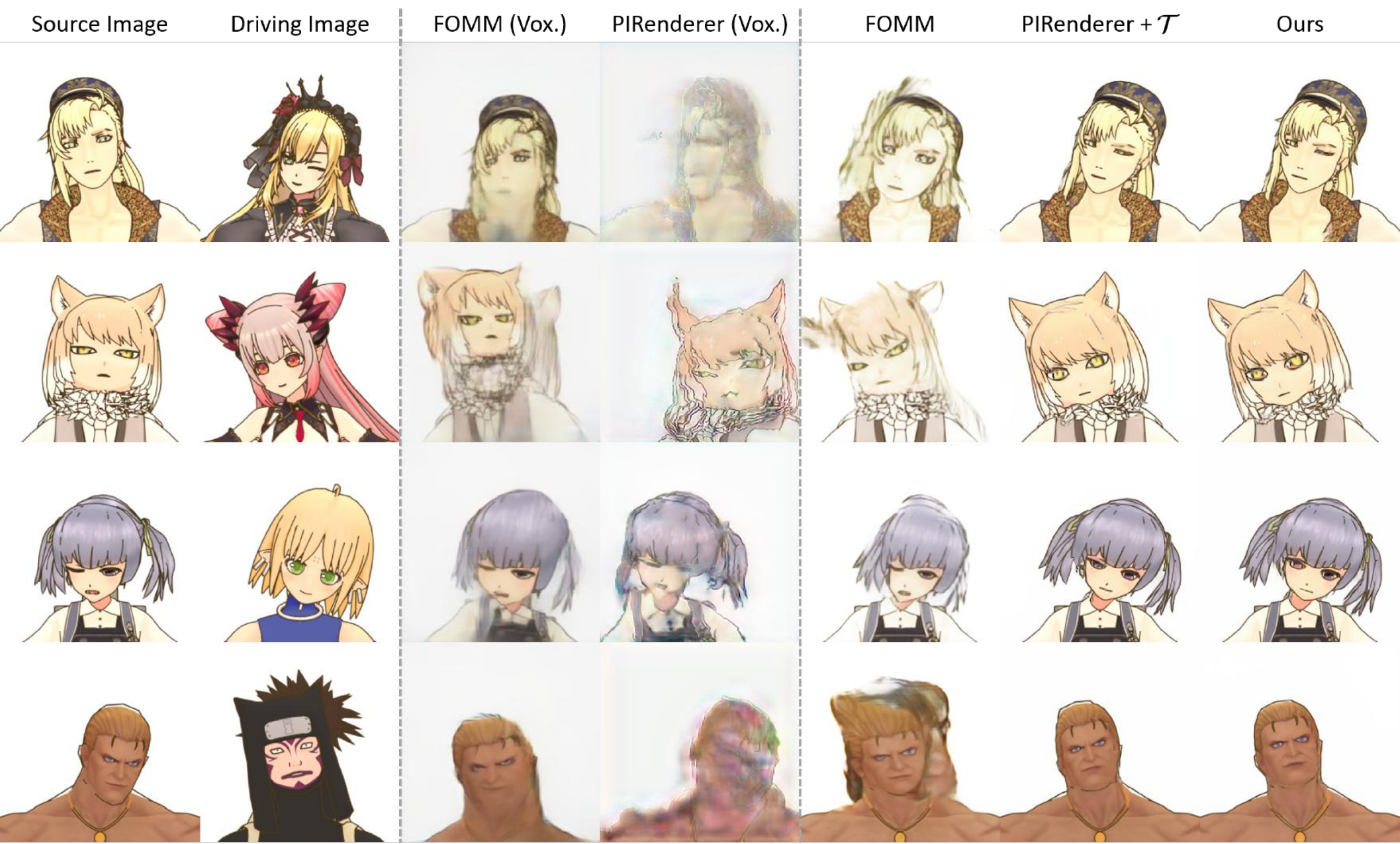}
    \caption{Qualitative comparison between our model and the baselines on cross-identity head reenactment given the images of the AnimeCeleb.}
    \label{supp-fig:baselines-cross-identity-anime}
\end{figure}

\begin{figure}[t!]
    \centering
    \includegraphics[width=1.0\linewidth]{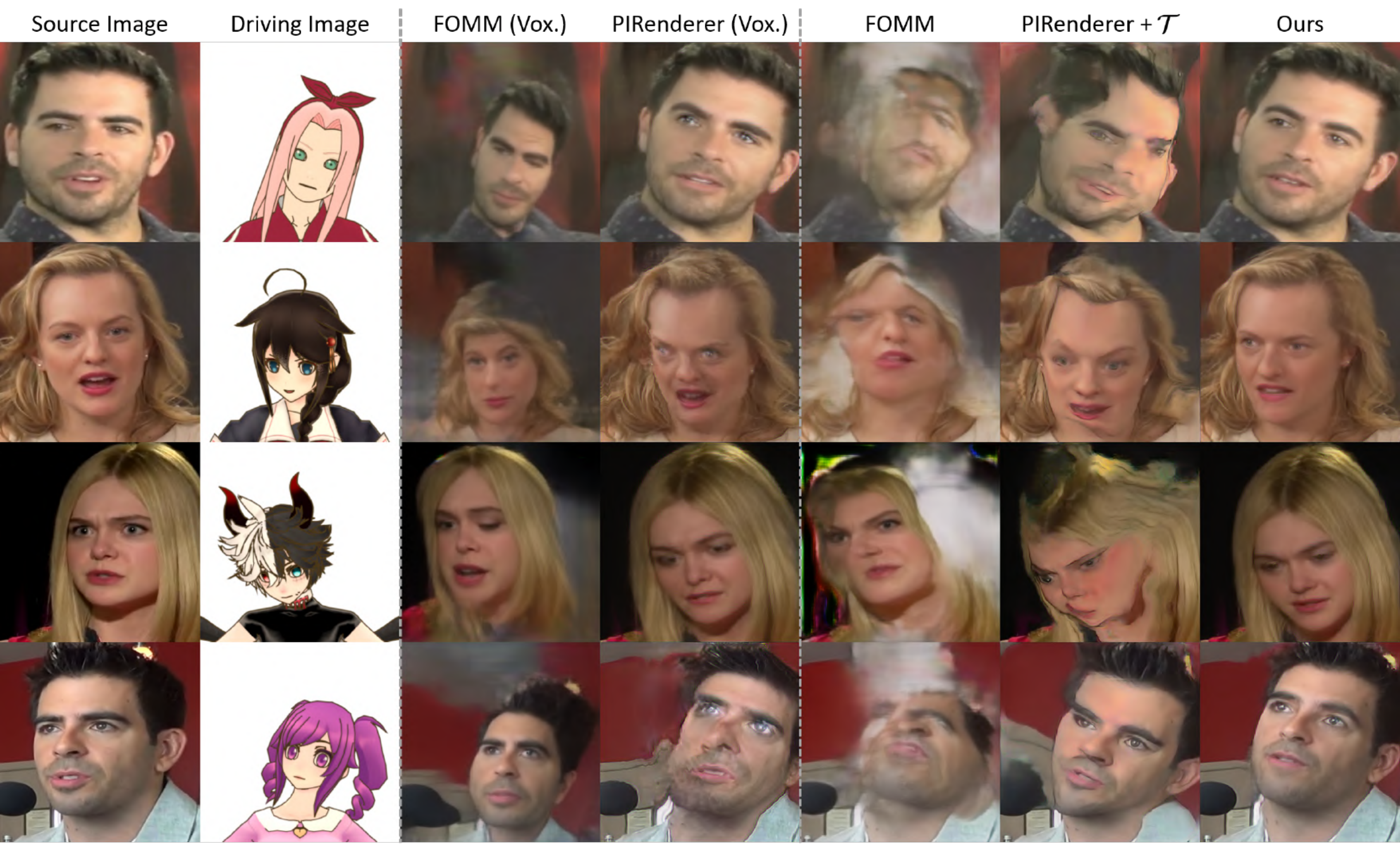}
    \caption{Qualitative comparison between our model and the baselines on cross-domain head reenactment given the source image from the VoxCeleb and the driving image from the AnimeCeleb (Anime. $\rightarrow$ Vox.).}
    \label{supp-fig:baselines-cross-domain-anime-vox}
\end{figure}

\begin{figure}[t!]
    \centering
    \includegraphics[width=1.0\linewidth]{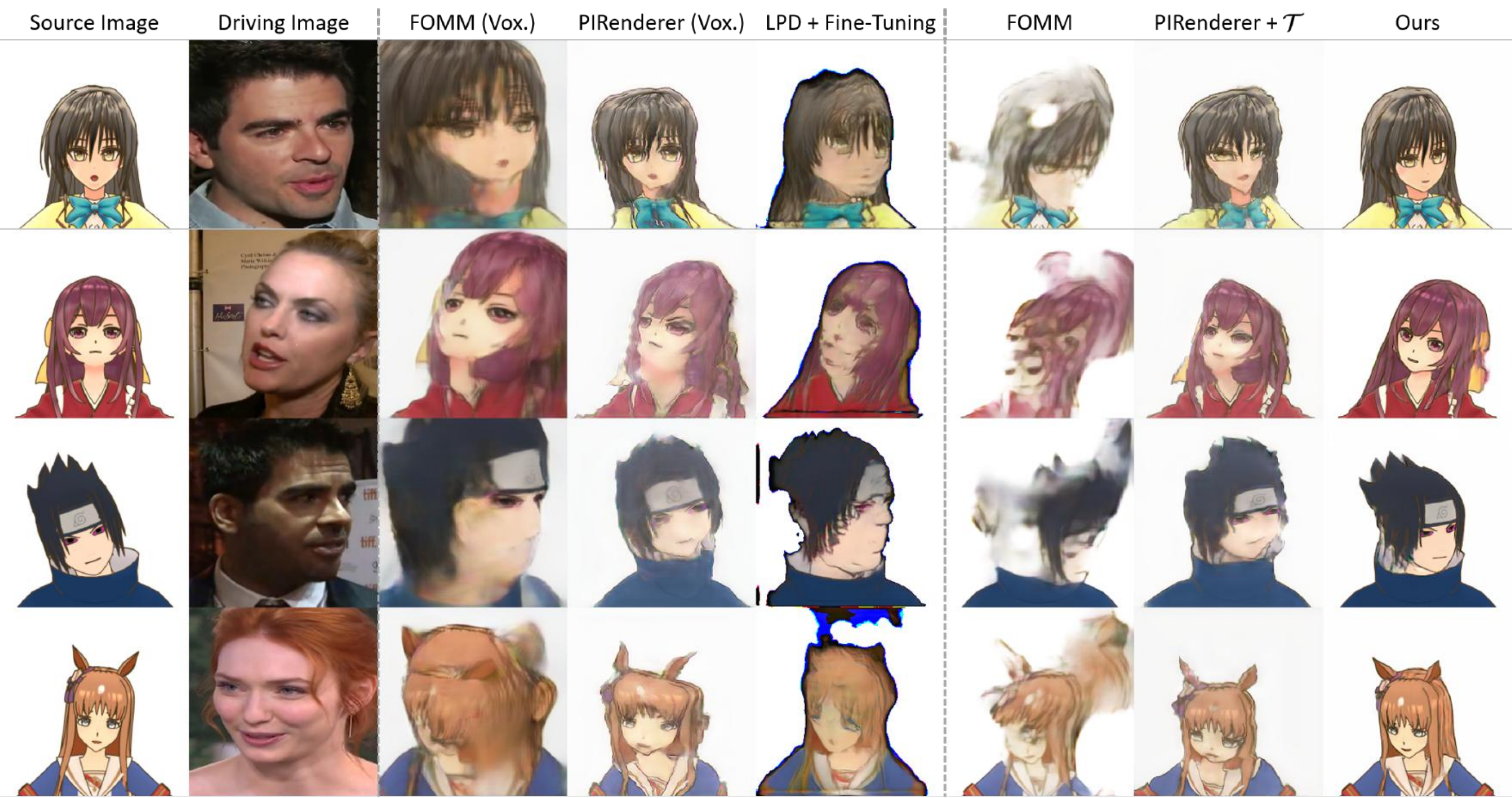}
    \caption{Qualitative comparison between our model and the baselines on cross-domain head reenactment given the source image from of the AnimeCeleb and the driving image from the VoxCeleb (Vox. $\rightarrow$ Anime.).}
    \label{supp-fig:baselines-cross-domain-vox-anime}
\end{figure}

Meanwhile, when performing on self-identity head reenactment using the AnimeCeleb, it is obvious that the models trained only with the VoxCeleb do not work well (the \textit{3rd} and the \textit{4th} columns in Fig.~\ref{supp-fig:baselines-self-identity-anime}).
In contrast, the models trained with the VoxCeleb and the AnimeCeleb show a promising performance (the \textit{6th} and the \textit{last} columns in Fig.~\ref{supp-fig:baselines-self-identity-anime}), yet the FOMM still has difficulty in synthesizing vivid textures of a source image (the \textit{5th} column in Fig.~\ref{supp-fig:baselines-self-identity-anime}).
In addition, Fig.~\ref{supp-fig:baselines-cross-identity-anime} shows similar results on cross-identity head reenactment, where the models trained with the VoxCeleb have performed poorly (the \textit{3th} and the \textit{4th} columns).
In contrast, the others trained with the AnimeCeleb successfully synthesize the outputs (the \textit{6th} and the \textit{last} columns\footnote{Note that both models use our pose mapping method.}) except for the FOMM (the \textit{5th} column). 
Furthermore, Fig.~\ref{supp-fig:baselines-cross-domain-anime-vox} and \ref{supp-fig:baselines-cross-domain-vox-anime} demonstrate that our model generates photo-realistic results compared to the baselines for cross domain head reenactment.

\begin{figure}[t!]
    \centering
    \includegraphics[width=1.0\linewidth]{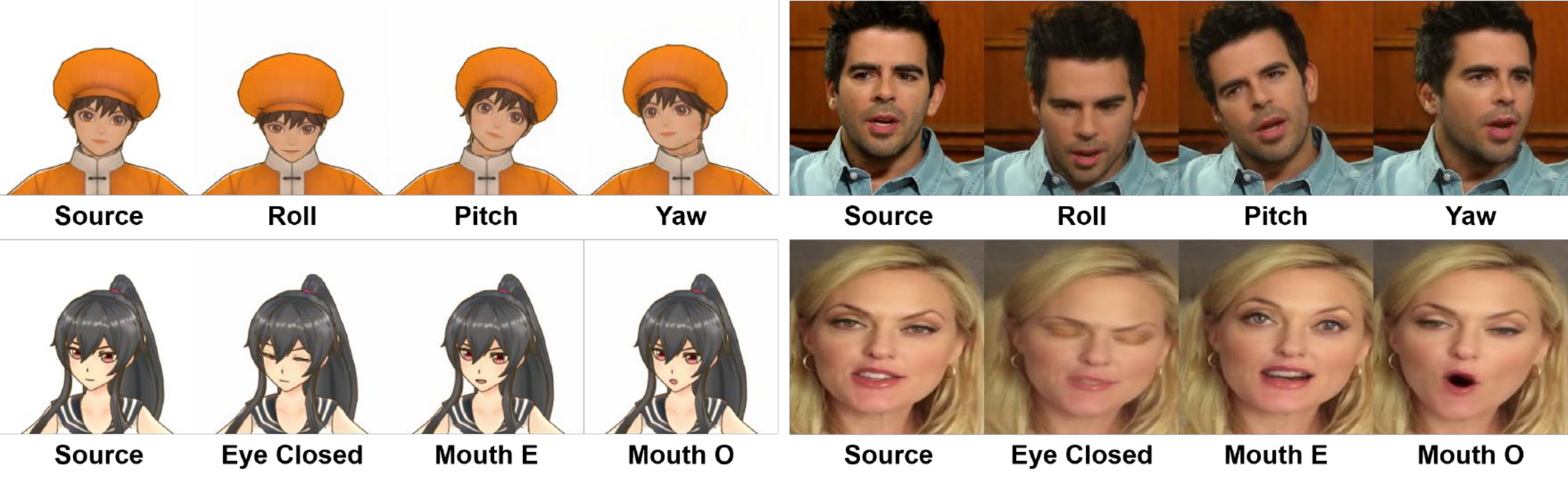}
    \vspace{-0.5cm}
    \caption{Intuitive image editing results on animation and human heads via controlling the semantics and the head angles.}
    \vspace{-0.5cm}
    \label{supp-fig:control_animo}
\end{figure}

\noindent\textbf{Intuitive Image Editing.}
One of the important applications of our model is to explicit control of a facial expression and head rotation on both the animation and human domain.
As shown in Fig.~\ref{supp-fig:control_animo}, the \model is capable of generating high-quality images steered by diverse semantics.
For example, an animation and human head can be controlled along roll, pitch and yaw axis (the \textit{1st} row in Fig.~\ref{supp-fig:control_animo}), and manipulating the facial expressions (\textit{i.e.}, eyes and a mouth) is achievable (the \textit{2nd} row in Fig.~\ref{supp-fig:control_animo}).

\begin{figure}[t!]
    \centering
    \includegraphics[width=0.9\linewidth]{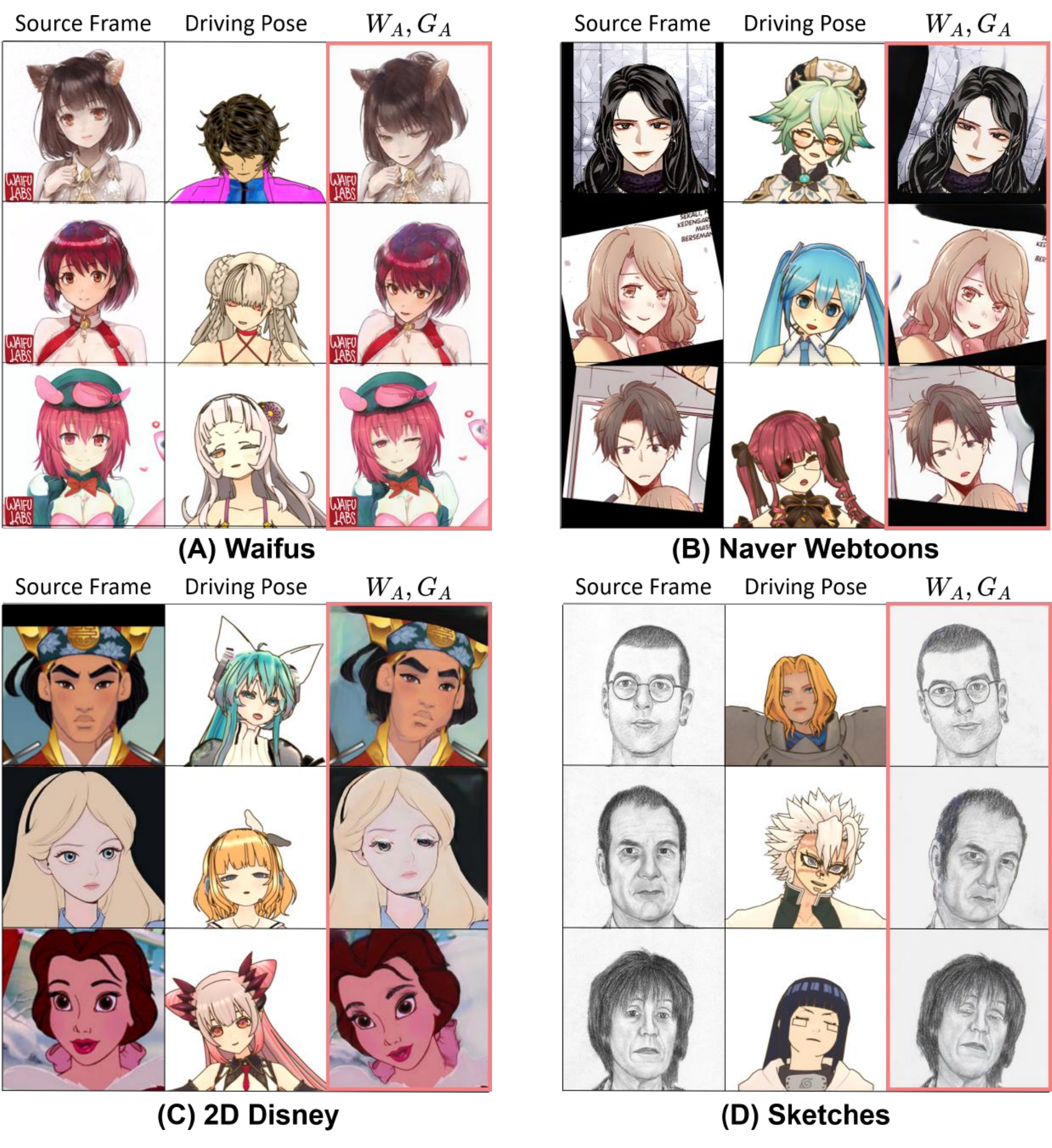}
    \vspace{-0.5cm}
    \caption{Additional head reenactment results on head images from various animation head samples.}
    \vspace{-0.5cm}
    \label{supp-fig:out-of-domain_animo}
\end{figure}

\noindent\textbf{Head Reenactment of Other Animation Images.}
% We present additional qualitative head reenactment results on various animation images in Fig.~\ref{supp-fig:out-of-domain_animo}.
In this experiment, we evaluate our model on multiple head image samples collected from different sources, including Waifu Labs, Naver Webtoon~\footnote{https://comic.naver.com/}, Face Sketches~\cite{ojha2021few}, 2D Disney~\footnote{https://toonify.photos/} as seen in Fig.~\ref{supp-fig:out-of-domain_animo}.
Given the trained $W_A$ and $G_A$ of the
\model, the poses of other animation images can be controlled with the guidance of driving poses.
However, we also find that there exist problems such as a background distortion and a lack of detailed expressions.
We discuss such problems in Section~\ref{discussions}.

% Figure - Animation Out-Domain Head Reenactment Results
\begin{figure}[t!]
    \centering
    \vspace{-0.3cm}
    \includegraphics[width=0.9\linewidth]{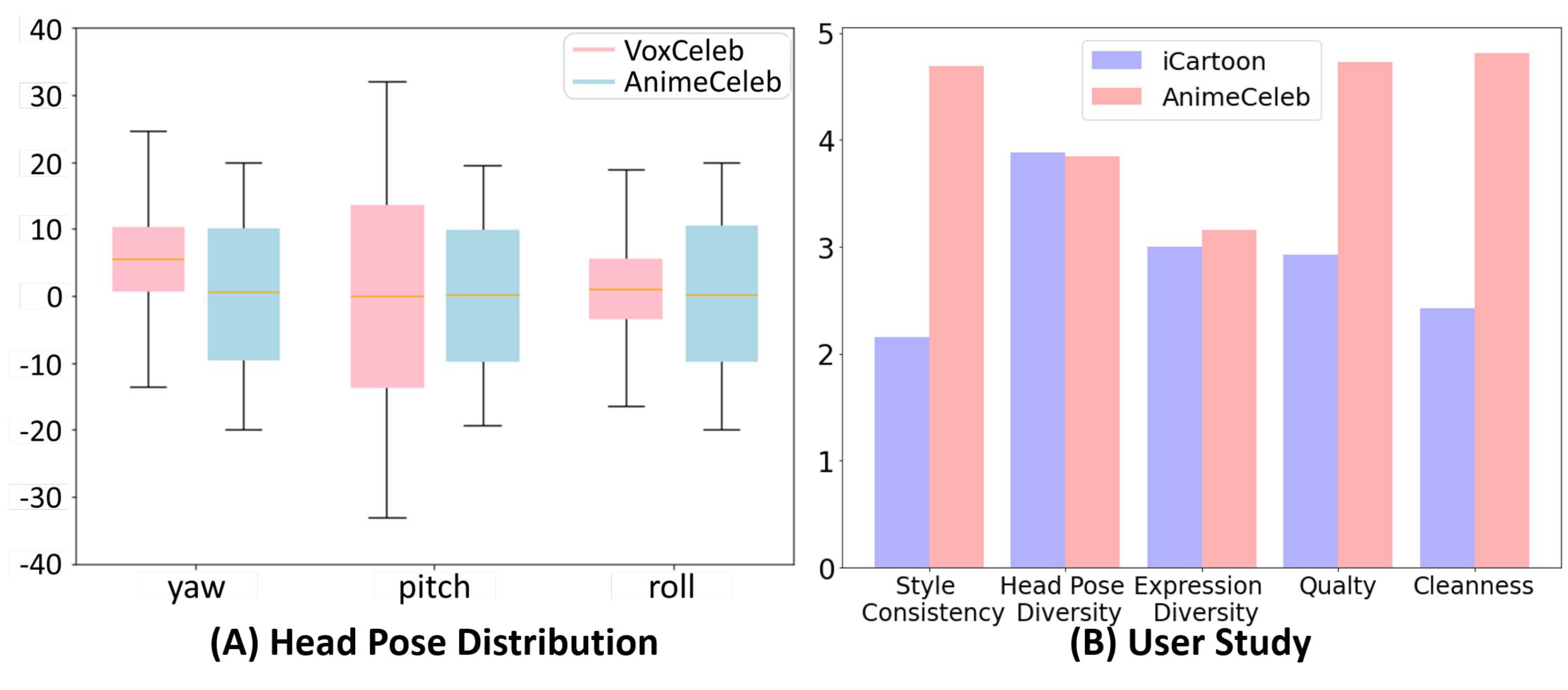}
    \vspace{-0.5cm}
    \caption{(A) Comparison of head pose statistics between VoxCeleb and AnimeCeleb. (B) User study results for comparison between iCartoon and AnimeCeleb. The higher score is better.}
    \vspace{-0.75cm}
    \label{supp-fig:userstudy}
\end{figure}

\noindent\textbf{Head Angle Comparison and User Study.}
Fig.~\ref{supp-fig:userstudy} (A) shows the ranges of head angles of 10K samples from each dataset. 
As can be seen, we determine the ranges of head poses in the scope of covering most samples of VoxCeleb.
For purpose of quantitative comparison with iCartoon, we conduct
a user study to compare the properties of datasets after see-
ing 100 samples from each dataset. 
As shown in Fig.~\ref{supp-fig:userstudy} (B), users positively evaluate the style consistency, quality\footnote{A low-resolution or defocused image is considered as low-quality one.} and cleanness\footnote{If a face is occluded with an object or incompletely cropped, then it is considered as a noisy image} of AnimeCeleb. 
Also, the users respond that AnimeCeleb has a comparable diversity of head pose and expression.

\clearpage

% Figure - Baselines Comparison (Self-Identity Vox -> Vox)

% Figure - Baselines Comparison (Cross-Identity Vox -> Vox)

% Figure - Baselines Comparison (Self-Identity Anime -> Anime)

% Figure - Baselines Comparison (Cross-Identity Anime -> Anime)

% Figure - Baselines Comparison (Cross-Identity Anime -> Vox)

% Figure - Baselines Comparison (Cross-Domain Vox -> Anime)

% Figure - Intuitive Image Editing

% Figure - Animation Out-Domain Head Reenactment Results

\clearpage

% Figure - High-resolution Image Samples
\begin{figure}[t!]
    \centering
    \includegraphics[width=0.95\linewidth]{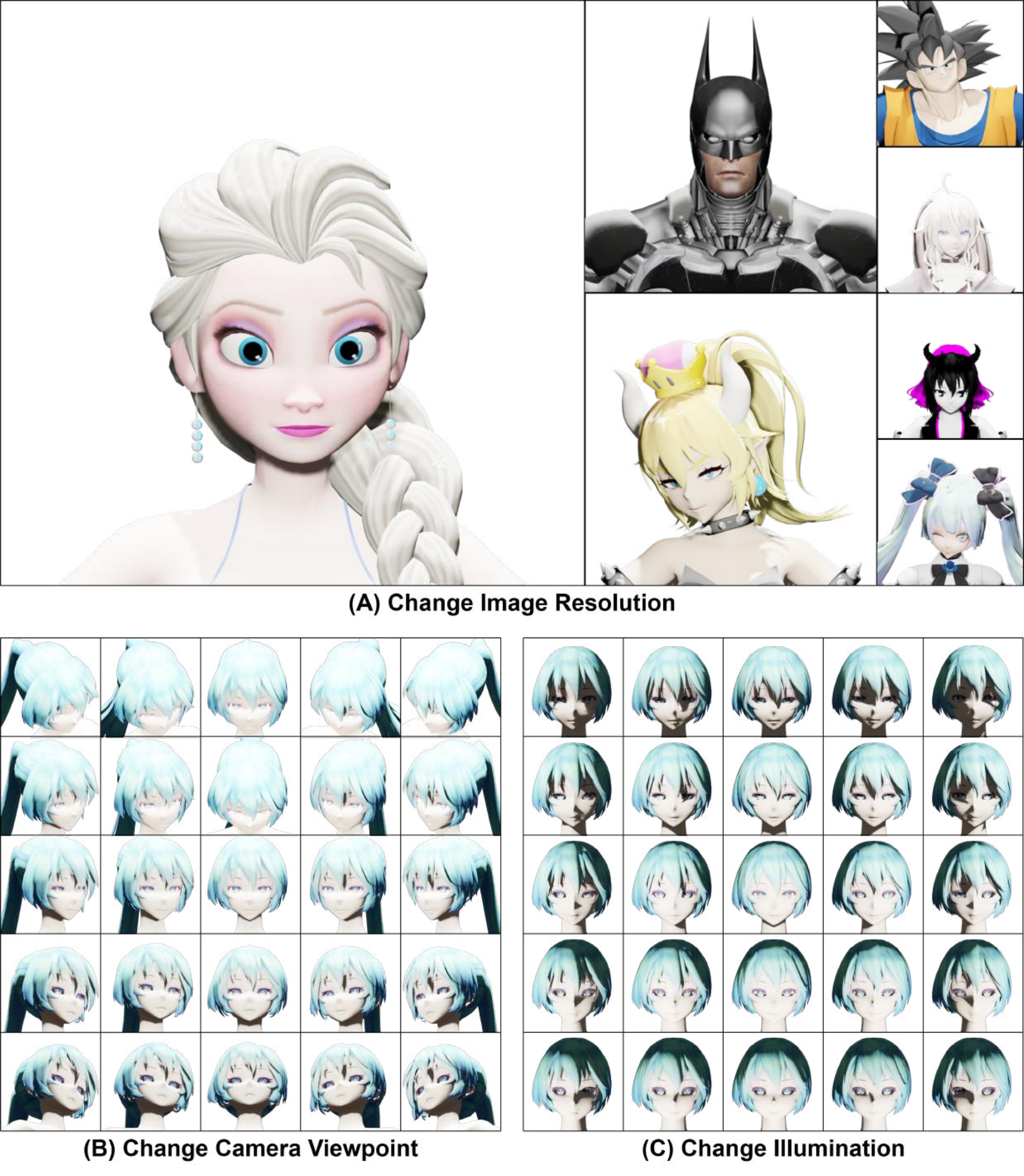}
    \vspace{-0.5cm}
    \caption{(A) Examples of rendered images with higher resolution (\textit{i.e.}, $1024\times1024$, $512\times512$, and $256\times256$) in order. 
    (B) We generate additional examples under different camera viewpoints from spherical coordinate system where the neck bone is the origin, ranging azimuth [-40\textdegree, 40\textdegree] and elevation [-40\textdegree, 40\textdegree].
    (C) Similar to (B) we render the images by relocating a light source position, ranging azimuth [-40\textdegree, 40\textdegree] and elevation [-40\textdegree, 40\textdegree] with setting the neck bone as the origin.}
    \vspace{-0.5cm}
    \label{supp-fig:megapixel}
\end{figure}

% Figure - Details
\begin{figure}[t!]
    \centering
    \includegraphics[width=0.95\linewidth]{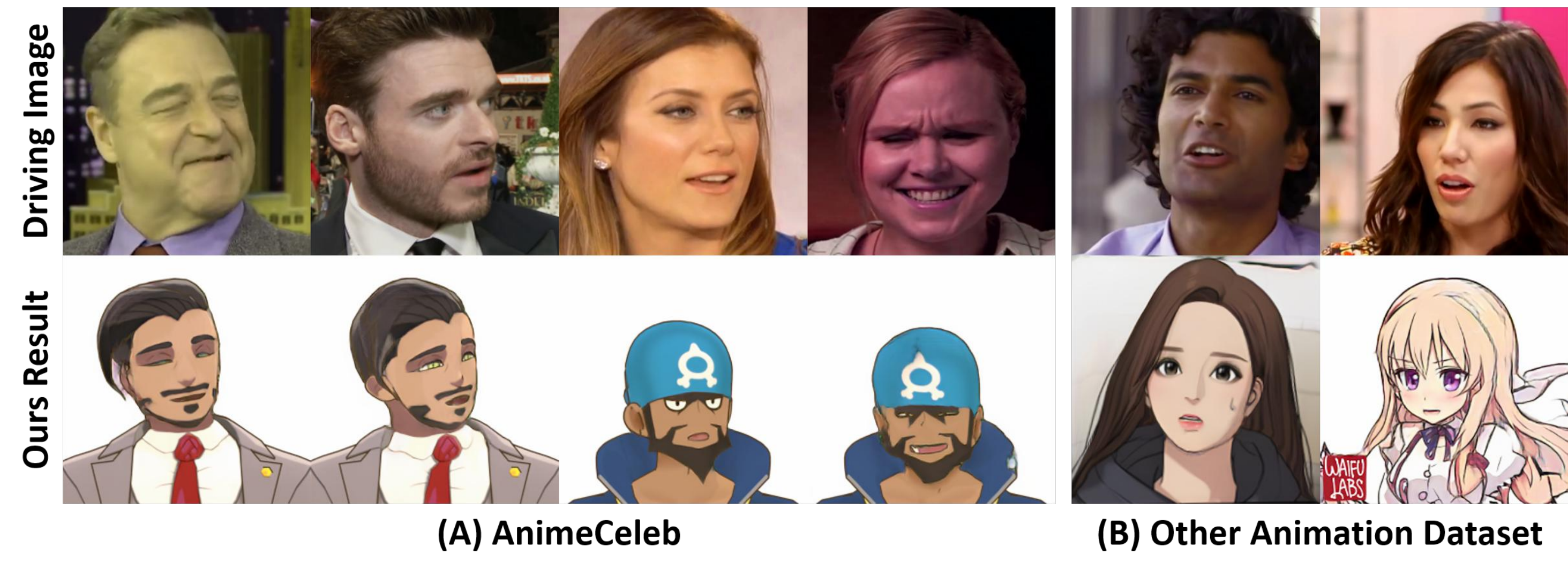}
    \caption{Additional cross-domain head reenactment results on (A) AnimeCeleb and (B) other animation datasets.}
    \label{supp-fig:cross-domain-detail}
\end{figure}

\section{Discussions}~\label{discussions}
In this section, we discuss potential issues and directions for improvement of the AnimeCeleb and the \model in further research.

\noindent\textbf{Extension of Creation Protocol.}
Due to the limited budget, the proposed pipeline is designed to generate a group of multi-pose yet single-view animation head images with the limited poses.
However, we believe that the AnimeCeleb has room for improvement in three aspects: (1) constructing high-quality images higher than $256\times256$, (2) obtaining multi-view animation head images by rotating the camera, and (3) building a various light-conditioned animation head dataset from changing the light source position. 
To prove these concepts, we present these samples in Fig.~\ref{supp-fig:megapixel}.
As seen in in Fig.~\ref{supp-fig:megapixel} (A), our data creation pipeline is able to render a higher resolution than $256\times256$ (\textit{e.g.}, $1024\times1024$).
This definitely allow us to construct a high-quality dataset in future research.
Next, the images of AnimeCeleb are created based on the frontal face, and thus do not span comprehensive appearances that can be created at various camera angles.
This is mainly due to the goal of the AnimeCeleb lies in constructing the public animation dataset, which is suitable for head reenactment.
A straightforward method to improve our creation process is to render an animation head at different camera angles in Blender as shown in Fig.~\ref{supp-fig:megapixel} (B).
Also, as can be seen in Fig.~\ref{supp-fig:megapixel} (C), we can control the illumination for the aim of generating animation head 
images under different light conditions.

\noindent\textbf{Diversity of the AnimeCeleb.}
One of the AnimeCeleb strengths lies in a wide spanning of animation characters.
However, we fixed the camera position with the aim of capturing frontal faces of animation characters during the AnimeCeleb generation process. 
Although this enables us to extract character face easily, the fixed camera position also constrained dataset diversity especially in terms of a translation.
In addition, we uniformly set a background of the generated image as 0 (\textit{i.e.}, white color).
Obviously, this weakens the capacity of a head reenactment model trained with the AnimeCeleb when handling a center-unaligned or complicated-background animation head image.
Our planned solution to these limitations is to develop a more flexible architecture that can consider translation parameters under this constraint.

\noindent\textbf{Limitations of the \model.}
We have found that when using 3DMM parameters obtained from the VoxCeleb, the \model often fails to reflect the detailed poses (\textit{e.g.}, eye or mouth pose).
Indeed, there are successful examples as shown in Fig.~\ref{supp-fig:cross-domain-detail}, our finding is that region sizes of lip and eyes are important to generate diverse images; more dynamics are tend to be entailed when a lip or eyes are noticeably large.
On the other hand, this is not the case when we use 3DMM parameters acquired by our pose mapping method with a pose vector from the AnimeCeleb.
We conclude that this behavior mainly stems from the fact that a pose from the VoxCeleb often does not identify the exact position of an eye or a mouth.
In future work, we will address this problem by considering expression detail correctness of the outputs during training.

In addition, since the images of the AnimeCeleb are center-aligned
and have no background, it is no surprise that there exists a performance degradation when an animation head image does not these conditions(\textit{e.g.}, containing complicated background).
To be specific, the generated outputs have an artifact at background and often loss the detailed poses (\textit{e.g.}, eye or mouth pose).
This behavior is also observed in previous studies~\cite{latentpose,firstorder,fewadvneural} when a position of a given head in an image is far from the training dataset distribution.
The solution to alleviate the problem by shifting a head position of a given source and driving image in the inference time~\footnote{https://github.com/shrubb/latent-pose-reenactment}.
Similar to these approaches, we plan to implement an additional preprocessing pipeline for an animation source image during the inference.

\end{document}